\documentclass[preprint,11pt]{elsarticle}
\usepackage{geometry}
\usepackage{amssymb}
\usepackage{subfigure}
\usepackage{longtable}
\usepackage{multirow}
\usepackage{array}
\usepackage{color}
\usepackage[ruled]{algorithm2e}
\usepackage{float}
\biboptions{numbers,sort&compress}
\usepackage{caption}
\usepackage{ulem}
\usepackage{epstopdf}
\usepackage{amsmath}
\usepackage{bm}
\usepackage{epstopdf}
\usepackage{cancel}  

\usepackage{ulem}
\usepackage{caption}

\newtheorem{definition}{Definition}

\journal{}

\begin{document}

\begin{frontmatter}

\title{Deep Asymmetric Hashing with Dual Semantic Regression and \\Class Structure Quantization}

\author[firstauthor,firstauthor2]{Jianglin Lu\corref{cor1}}
\author[firstauthor3]{Hailing Wang\corref{cor1}}
\author[firstauthor,firstauthor2]{Jie Zhou\corref{cor2}}
\author[firstauthor4]{Mengfan Yan}
\author[firstauthor]{Jiajun Wen}

\cortext[cor1]{Equal contribution.}
\cortext[cor2]{Corresponding author. E-mail addresses: jie\_jpu@163.com (J. Zhou).}

\address[firstauthor]{College of Computer Science and Software Engineering, Shenzhen University, \\ Shenzhen 518060, China}
\address[firstauthor2]{SZU Branch, Shenzhen Institute of Artificial Intelligence and Robotics for Society,\\ Shenzhen 518060, China}
\address[firstauthor3]{College of Computer Software, Tianjin University, Tianjin 300350, China}
\address[firstauthor4]{College of Engineering and Computer Science, Australian National University, \\ Canberra ACT 0200, Australia}

\begin{abstract}
Recently, deep hashing methods have been widely used in image retrieval task.
Most existing deep hashing approaches adopt one-to-one quantization to reduce information loss. However, such \textit{class-unrelated quantization} cannot give discriminative feedback for network training. In addition, these methods only utilize single label to integrate supervision information of data for hashing function learning, which may result in inferior network generalization performance and relatively low-quality hash codes since the inter-class information of data is totally ignored. 
In this paper, we propose a dual semantic asymmetric hashing (DSAH) method, which generates discriminative hash codes under three-fold constraints. Firstly, DSAH utilizes class prior to conduct class structure quantization so as to transmit class information during the quantization process. Secondly, a simple yet effective label mechanism is designed to characterize both the intra-class compactness and inter-class separability of data, thereby achieving semantic-sensitive binary code learning. Finally, a meaningful pairwise similarity preserving loss is devised to minimize the distances between class-related network outputs based on an affinity graph. With these three main components, high-quality hash codes can be generated through network. Extensive experiments conducted on various data sets demonstrate the superiority of DSAH in comparison with state-of-the-art deep hashing methods.

\end{abstract}


\end{frontmatter}

\section{Introduction}

Hashing has received extensive attention over the past few years in the field of image search. It aims to transform high-dimensional data into compact binary representations, while maintaining original similarity structure of data \cite{48}. 
Owing to its low storage consumption and high computation efficiency, hashing has been successfully applied in many fields, such as deep neural network quantization \cite{5}, sketch retrieval \cite{54}, domain adaptation \cite{47}, dimensionality reduction \cite{18}, and multi-label image search \cite{43}.

Traditional hashing methods can be roughly divided into two categories, including data-independent and data-dependent methods. The typical data-independent hashing method, locality sensitive hashing (LSH) \cite{8}, generates binary codes of data via random projections. In \cite{30}, a global low-density locality sensitive hashing (GLDH) is proposed to produce a low error hyperplane partition and find the global low-density hyperplanes. Although these methods achieve good retrieval efficiency, they require relatively long hash codes to obtain high precision and recall rates \cite{42}. In order to learn compact binary codes, a family of data-dependent hashing methods have been proposed, which can be further classified into unsupervised and supervised methods. Most of unsupervised hashing methods aim to learn compact binary codes embedded on their intrinsic manifolds \cite{31, 42, 51, 49}. These methods typically construct a graph incorporating affinity relationship of data for effective binary code learning. For instance, spectral hashing (SH) \cite{51} aims to preserve the intrinsic manifold structure of data in the Hamming space. Due to the difficulty in optimization, SH adopts relaxation strategy to optimize its objective function by removing the binary constraint directly. However, such relaxation strategy makes a great information loss, resulting in performance degradation. 
To this end, discrete graph hashing (DGH) \cite{34} explicitly handles the binary constraint and directly learns hash codes, which better characterizes the manifold structure of data and produces higher-quality binary codes. Being different from DGH, iterative quantization (ITQ) \cite{10} attempts to find an orthogonal rotation matrix to minimize the information loss between hash codes and data representations.
The binary reconstructive embedding (BRE) \cite{21} explicitly minimizes the reconstruction error between the original distances and the Hamming distances of the corresponding hash codes, which avoids the restrictions of data distribution assumptions. Another potential problem in SH is that the construction of similarity matrix is prohibitively time-consuming, limiting its application in large-scale data sets. To overcome this drawback, some efficient graph construction hashing methods \cite{15, 35} have been proposed, which construct an anchor graph with asymmetric similarity metric.    
Nevertheless, unsupervised hashing methods fail to achieve satisfactory performance due to the lack of semantic information, hindering their applications in practice.

As indicated in \cite{59}, high-level semantic description of an object often differs from the extracted low-level feature descriptors. To make full use of label information of data, various supervised hashing methods have been proposed \cite{13, 14, 41}, which achieve superior retrieval performance than unsupervised methods. These methods attempt to learn semantic-sensitivity hash codes guided by manually-annotated labels. For example, supervised discrete hashing (SDH) \cite{41} aims to generate the optimal binary codes for linear classification, which keeps the discrete constraint in optimization and utilizes full semantic information for discriminative binary code learning. To enlarge the regression boundary and exploit more precise regression target, supervised discrete hashing with relaxation (SDHR) \cite{14} learns the discriminative hash codes and regression targets simultaneously in a joint optimization objective. Being different from these methods which regress binary codes into data labels, fast supervised discrete hashing (FSDH) \cite{13} regresses the data labels into binary codes, achieving faster training speed and better retrieval performance. More recently, the supervised robust discrete multimodal hashing \cite{25} and supervised discrete multi-view hashing \cite{37} have been proposed, which extend SDH into the cross-media scenario and FSDH into the multi-view application, respectively. The above-mentioned supervised hashing methods can obtain optimal hash codes that are penalized by a discrete constraint. Nevertheless, these methods usually encode images by hand-crafted features. As a result, they lack the feature learning ability to generate high-quality binary codes.

Inspired by great advances in deep learning, many deep hashing methods \cite{6, 19, 26, 27, 28, 40} have been proposed to perform feature learning and hashing function learning simultaneously in an end-to-end framework. The representative methods include deep pairwise supervised hashing (DPSH) \cite{28}, deep asymmetric pairwise hashing (DAPH) \cite{40}, deep supervised discrete hashing (DSDH) \cite{27}, deep joint semantic-embedding hashing (DSEH) \cite{26}, and deep anchor graph hashing (DAGH) \cite{6}. 
Generally speaking, these methods utilize a negative log-likelihood-based similarity preserving loss to learn deep hashing functions. However, such loss faces the problem of imbalance data since there are far fewer similar pairs than dissimilar pairs \cite{3}. 
In addition to this limitation, there are two more potential drawbacks in existing deep hashing methods.    
Firstly, due to the non-differentiable sign function, deep hashing methods directly remove the discrete constraint during network training, and obtain real-valued network outputs as the approximate binary codes. In oder to reduce information loss, they typically penalize the distance between each real-valued embedding and its corresponding binary code (which is generally referred to as quantization).  Nevertheless, such one-to-one quantization strategy only consider the quantization error for each individual sample rather than class-related samples, leading to a \textit{class-unrelated quantization} problem. 
Secondly, in the process of binary code learning, these methods put undue emphasis on intra-class compactness while ignoring inter-class separability of data. As a result, the semantic knowledge of data is not fully utilized to learn semantic-sensitive binary codes.

\begin{figure}[!tb]
	\centering
	\includegraphics[scale=1.5,trim=200 690  225 30,clip]{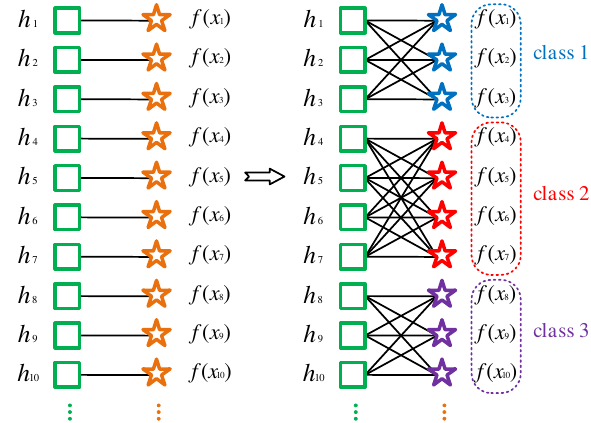}
	\caption{Illustration of class structure quantization. Left: one-to-one quantization. Existing deep hashing methods minimize the distance between binary code $\mathbf{h}_i$ and its corresponding real-valued embedding $f(\mathbf{x}_i)$ for each individual sample $\mathbf{x}_i$. Right: class structure quantization. The proposed method minimizes the quantization errors between binary code $\mathbf{h}_i$ and all its class-related real-valued embeddings $f(\mathbf{x})$.} \label{class_structure_quantization}
\end{figure}

Motivated by these issues, we attempt to develop a novel quantization strategy to minimize the distances between each binary code and all its class-related real-valued embeddings (as illustrated in Figure \ref{class_structure_quantization}).
Additionally, we aim to design an effective label mechanism to characterize both the intra-class and inter-class information of data. We hope that the proposed method can make the binary codes of the same class to be similar and the binary codes of different classes far away from each other. 
To achieve these goals, this paper proposes a novel deep supervised hashing framework called dual semantic asymmetric hashing (DSAH), which jointly conducts dual semantic regression to integrate both the intra-class and inter-class information of data, and class structure quantization to minimize the gap between each hash code and all its class-related embeddings. Being different from previous methods which use likelihood-based similarity preserving loss, DSAH adopts a novel similarity preserving loss to utilize semantic knowledge of data. Specifically, this loss minimizes the distances between class-related real-valued embeddings via an affinity graph that is pre-constructed from label information of data.

The main contributions of this paper can be summarized as follows:

\begin{itemize}
	\item We propose a novel deep supervised hashing framework called DSAH, which adopts an interesting similarity preserving strategy based on distance-similarity product function. Such strategy minimizes the product of data distance calculated from Hamming space and data similarity derived from input space.
	
	\item DSAH considers the quantization errors between each hash code and all its class-related real-valued embeddings. It subtly incorporates class prior of data into the quantization stage, thereby providing discriminative feedback for network training.
	
	\item To learn semantic-sensitive binary codes, a dual semantic regression loss supported by a simple yet effective label mechanism is well designed to characterize the intra-class compactness and inter-class separability of data. 
	
	\item Extensive experiments show that the proposed method provides much superior generalization of network than existing state-of-the-art deep hashing methods.	
\end{itemize}

The rest of this paper is organized as follows. The related works are briefly introduced in Section 2. Section 3 shows the framework of the proposed DSAH and provides the corresponding optimization algorithm. Extensive experiments are presented in Section 4. Section 5 gives the conclusions and our future work.

\begin{table}[!tb]
	\begin{center}
		\small
		\setlength{\tabcolsep}{0.8mm}
		\caption{The list of notations.}
		\label{notations}
		\begin{tabular}{l|l}	
			\hline	
			\hline
			Notation  &Description  \\
			\hline	
			$\mathbf{X}\in \Re^{n\times d}$  & the matrix of training data  \\
			$\mathbf{x}_i\in \Re^{d}$  & the $i$-row of $\mathbf{X}$   \\
			$\mathbf{X}_{ij}\in \Re$  & the element in the $i$-row and $j$-column of $\mathbf{X}$   \\
			$\mathbf{Y}\in \Re^{n\times k}$  & the intra-class label matrix  \\	
			$\mathbf{R}\in \Re^{n\times k}$  & the inter-class label matrix  \\	
			$\mathbf{H}\in \Re^{n\times c}$  & the binary code matrix of training data $\mathbf{X}$\\
			$\mathbf{h}_i\in \Re^{c}$ &the binary code of the training sample $\mathbf{x}_i$ \\
			$\mathbf{U}\in \Re^{m\times c}$  &the outputs of the first network  \\
			$\mathbf{V}\in \Re^{m\times c}$  &the outputs of the second network \\			
			$\mathbf{M}_1\in \Re^{k\times c}$ &the intra-class regression matrix  \\
			$\mathbf{M}_2\in \Re^{k\times c}$ &the inter-class  regression matrix  \\
			$\mathbf{S}\in \Re^{n\times m}$ &the indicator matrix   \\
			$\mathbf{W}\in \Re^{m\times m}$& the pairwise similarity matrix  \\
			$\widetilde{\mathbf{W}}\in \Re^{m\times m}$& the nonnegative pairwise similarity matrix  \\
			$\mathbf{A}\in \Re^{n\times n}$ &the pairwise label matrix  \\
			$n$  & the number of training samples\\
			$d$  & the dimension of training samples\\
			$m$  & the number of training samples used in network \\	
			$k$  & the number of classes   \\
			$c$  & the length of binary code  \\		
			\hline
			\hline
		\end{tabular}
	\end{center}
\end{table}

\section{Related Works}
In this section, we briefly review some related hashing approaches, including supervised hashing, deep hashing, and asymmetric hashing. The notions used in this paper and their corresponding descriptions are listed in Table \ref{notations} for ease of reading.

\subsection{Supervised Hashing with Discrete Constraint}  
As mentioned above, supervised hashing methods \cite{13,37,41} aim to learn discriminative hash codes guided by manually-annotated labels. The most representative method is supervised discrete hashing (SDH) \cite{41}, which generates the optimal binary codes for linear classification. Given the training data $\mathbf{X}\in \Re^{n \times d}$ and the corresponding label $\mathbf{Y}\in \{0,1\}^{n \times k}$ (where $n$, $d$, $k$ are the number of samples, the dimensionality of data, and the number of classes, respectively), SDH aims to optimize the following objective function:
\begin{align}
\min_{\mathbf{H},\mathbf{M},\mathbf{F}}||\mathbf{Y}-\mathbf{HM}^T||_F^2+\lambda ||\mathbf{M}||_F^2+\nu||\mathbf{H}-\mathbf{F}(\mathbf{X})||_F^2 \label{SDH_Obj} 
\end{align}
where $\lambda$ and $\nu$ are two penalty parameters, $\mathbf{H}\in \{-1,1\}^{n \times c}$ is the binary code matrix of $\mathbf{X}$, $c$ is the length of binary codes, and $\mathbf{M}\in \Re^{k \times c}$ is a regression matrix. The last term $||\mathbf{H}-\mathbf{F}(\mathbf{X})||_F^2$ models the fitting errors between the binary codes $\mathbf{H}$ and continuous embeddings $\mathbf{F}(\mathbf{X})$. In optimization process, SDH keeps the discrete constraint and directly learns the binary codes, rather than using the relaxation strategy. Being different from SDH, fast supervised discrete hashing (FSDH) \cite{13} is proposed to regress data labels into binary codes, achieving faster training speed and better retrieval performance. 
However, since these methods only use single label for regression, they ignore the inter-class separability of data. As a result, the semantic knowledge of data is not fully utilized to learn semantic-sensitive hash codes.

\subsection{Deep Hashing with One-to-One Quantization}
Deep hashing methods \cite{16, 19, 26, 27, 28, 40, 55} aim to perform feature learning and hashing function learning jointly in an end-to-end framework. 
Generally speaking, these methods utilize the following negative log-likelihood-based similarity preserving loss to learn deep hashing functions:
\begin{equation}
\min_{\mathbf{H}}\sum_{\mathbf{A}_{ij}\in \mathbf{A}}(\mathtt{log} (1+e^{\mathbf{\Theta}_{ij}})-\mathbf{A}_{ij}{\mathbf{\Theta}_{ij}}) \label{deephashing2}
\end{equation}
where $\mathbf{\Theta}_{ij} = \frac{1}{2}\mathbf{h}_i^T\mathbf{h}_j$, $\mathbf{h}_i$ is the $i$-th row of $\mathbf{H}$, and $\mathbf{A}\in \Re^{n\times n}$ denotes a pairwise label matrix. 
If $\mathbf{x}_i$ and $\mathbf{x}_j$ are similar, $\mathbf{A}_{ij}=1$, otherwise, $\mathbf{A}_{ij}=0$. 
In the training process, these methods directly remove the discrete constraint  and obtain real-valued network outputs as the approximate binary codes. To reduce the information loss caused by real-valued approximation, these methods typically add a one-to-one quantization regularization to penalize the distance between each real-valued network output and its corresponding hash code as follows:
\begin{equation}
\min_{\mathbf{H}}\sum_{\mathbf{A}_{ij}\in \mathbf{A}}(\mathtt{log} (1+e^{\widetilde{\mathbf{\Theta}}_{ij}})-\mathbf{A}_{ij}{\widetilde{\mathbf{\Theta}}_{ij}})+\gamma\sum_{i}||\mathbf{u}_i-\mathbf{h}_i||^2 \label{deephashing3} 
\end{equation}
where $\gamma$ is a penalty parameter, $\widetilde{\mathbf{\Theta}}_{ij} = \frac{1}{2}\mathbf{u}_i^T\mathbf{u}_j$, and $\mathbf{u}_i \in \Re^{c}$ is the network output of data $\mathbf{x}_i$.
As we can see, these methods perform \textit{class-unrelated quantization} since they minimize the quantization error for each individual sample rather than class-related samples. A potential drawback is that such quantization strategy ignores the class prior of data and thus cannot give discriminative feedback for network training.

\subsection{Asymmetric Hashing}
Some early research works on asymmetric hashing \cite{9, 11} employ asymmetric distance metric for similarity preserving, where the binary codes of training and test sets are generated from the same hashing function that needs to be learned. This paper focuses on functional space, where the distribution gaps between training and test samples are characterized by two distinct hashing functions. 
As indicated in \cite{39}, using asymmetric hashing mechanism can perform better approximation of target similarity with shorter code lengths. Inspired by this, some deep asymmetric hashing approaches have been proposed, such as deep asymmetric pairwise hashing (DAPH) \cite{40}, asymmetric deep supervised hashing (ADSH) \cite{20}, nonlinear asymmetric multi-valued hashing \cite{7}, and multi-head asymmetric hashing \cite{38}. 
Technically, these methods approximate the similarity between samples $\mathbf{x}_i$ and $\mathbf{x}_j$ as the Hamming distance between $p(\mathbf{x}_i)$ and $q(\mathbf{x}_j)$, where $p(\cdot)$ and $q(\cdot)$ are two distinct hashing functions. The more similar the two samples $\mathbf{x}_i$ and $\mathbf{x}_j$ are, the smaller the Hamming distance between $p(\mathbf{x}_i)$ and $q(\mathbf{x}_j)$ should be.

\begin{figure}[!tb]
	\centering
	\includegraphics[scale=0.48,trim=10 170 0 20,clip]{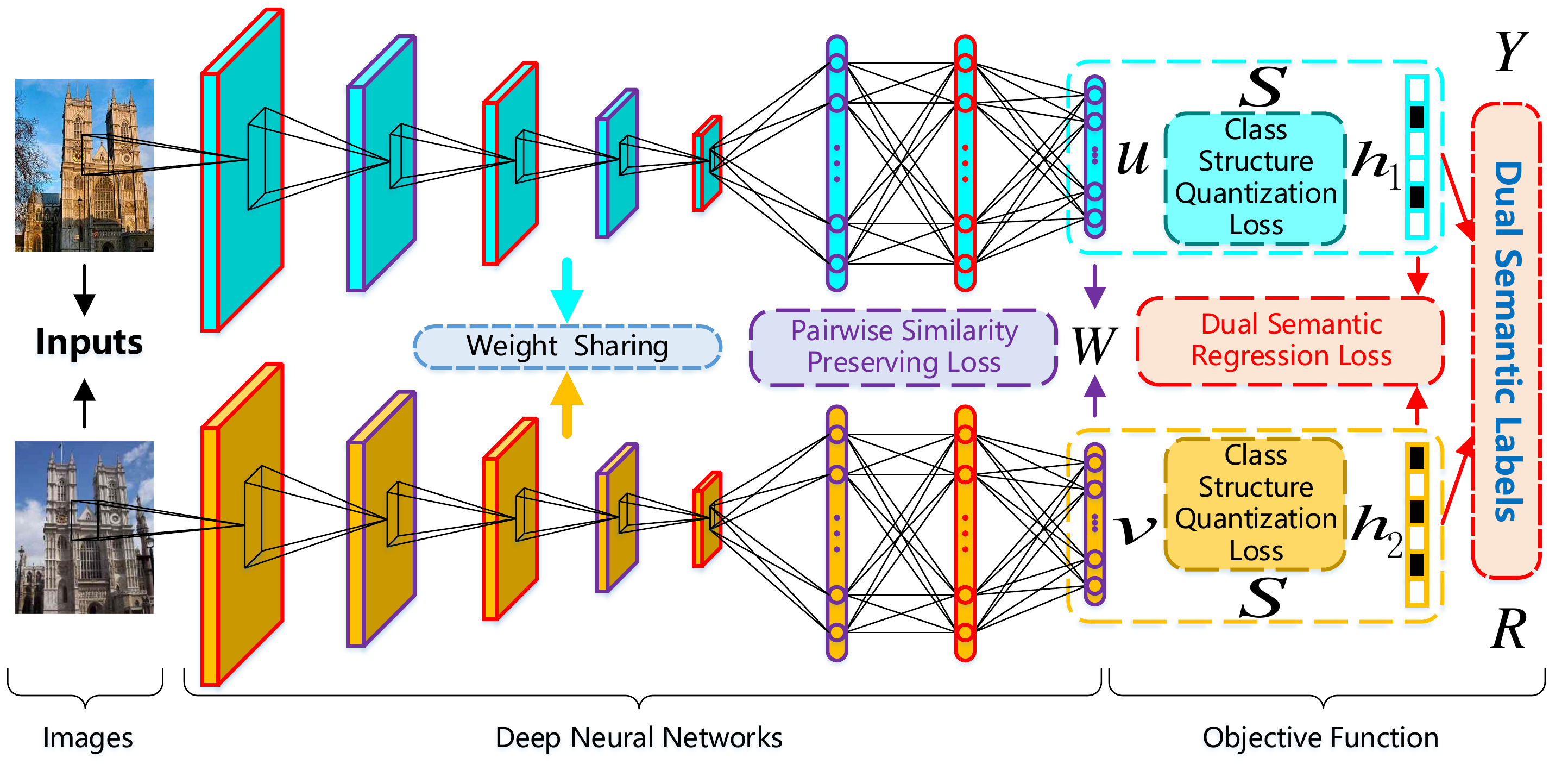}
	\caption{The end-to-end deep architecture of the proposed DSAH.} 
\end{figure}

\section{Dual Semantic Asymmetric Hashing}
This sections gives the overall framework of the proposed DSAH and designs an iterative algorithm to optimize its objective function. Then, we show how to solve the out-of-sample extension problem and provide an accelerated version for the proposed DSAH.
\subsection{Overall Framework}
Figure 2 gives an illustration of the proposed DSAH framework. As shown in Figure 2, DSAH mainly contains three parts, including dual semantic regression $\mathcal{R}$, pairwise similarity preserving $\mathcal{P}$, and class structure quantization $\mathcal{Q}$. The overall framework of the proposed DSAH can be formulated as follows:
\begin{align}
&\min_\mathbf{H}\mathcal{J}=\mathcal{R}+\alpha_1\mathcal{P}+\alpha_2\mathcal{Q} 
\qquad \mathrm{s.t.}\  \mathbf{H} \in \{-1,1\}^{n \times c},\ \mathbf{H}^T\mathbf{1}=0
 \label{obj}
\end{align}
where $\alpha_1$ and $\alpha_2$ are two nonnegative parameters.
The balance constraint $\mathbf{H}^T\mathbf{1}=0$ is added to make each bit have equal chance to be $1$ or $-1$.
The following shows how we design such three components of the proposed method.

\subsection{Dual Semantic Regression}
In order to exploit both intra-class and inter-class information of data, we first define a simple yet effective label mechanism for discriminative binary code learning.

\begin{definition}
	If the $i$-th target belongs to $j$-th class, $\mathbf{Y}_{ij}=\sqrt{\beta_1}$ and $\mathbf{R}_{ij}=0$. Otherwise, $\mathbf{Y}_{ij}=0$ and $\mathbf{R}_{ij}=\sqrt{\beta_2}$. The pairwise labels $\mathbf{Y}$ and $\mathbf{R}$ are called dual semantic labels, where $\beta_1$ and $\beta_2$ are two nonnegative hyper-parameters. 
\end{definition}

With the defined dual semantic labels $\mathbf{Y}$ and $\mathbf{R}$, we develop the following dual semantic regression loss $\mathcal{R}$ to learn semantic-sensitive binary codes:

\begin{equation}
\min_{\mathbf{H},\mathbf{M}_1,\mathbf{M}_2}\mathcal{R}=\underbrace{||\sqrt{\beta_1}\mathbf{H}-\mathbf{YM}_1||_F^2}_{intra-class\ loss}-\underbrace{||\sqrt{\beta_2}\mathbf{H}-\mathbf{RM}_2||_F^2}_{inter-class\ loss} \label{Regression}
\end{equation}
where $\mathbf{M}_1\in \Re^{k\times c}$ and $\mathbf{M}_2\in \Re^{k\times c}$ are two regression matrices.
Differing from \cite{13, 27, 41}, the proposed dual semantic regression $\mathcal{R}$ characterizes the intra-class compactness and inter-class separability of data. The \textit{intra-class loss} penalizes intra-class classification errors while the \textit{inter-class loss} integrates inter-class information of data. 
In fact, the nonnegative hyper-parameters $\beta_1$ and $\beta_2$ control the weight between \textit{intra-class loss} and \textit{inter-class loss}. Suppose that $\mathbf{Y}=\sqrt{\beta_1}\mathbf{\hat{Y}}$ and $\mathbf{R}=\sqrt{\beta_2}\mathbf{\hat{R}}$, where $\mathbf{\hat{Y}}$ and $\mathbf{\hat{R}}$ are binary matrices with elements $1$ and $0$. Then, the dual semantic regression $\mathcal{R}$ is equivalent to the following optimization objective:
\begin{equation}
\min_{\mathbf{H},\mathbf{M}_1,\mathbf{M}_2}\beta_1||\mathbf{H}-\mathbf{\hat{Y}M}_1||_F^2-\beta_2||\mathbf{H}-\mathbf{\hat{R}M}_2||_F^2  
\end{equation}
Obviously, the hyper-parameters $\beta_1$ and $\beta_2$ can adjust the proportion of intra-class and inter-class information of data.

\subsection{Pairwise Similarity Preserving}
Based on the pairwise labels of data, existing deep hashing methods generally utilize the following pairwise similarity preserving loss for hashing function learning:
\begin{align}
&\min_{p,q}\sum_{i}\sum_{j}\varPi\left(\ (p_i,q_j)\  | \ \mathbf{W}_{ij}\ \right)
\end{align} 
where $p$ and $q$ are two hashing functions, $p_i=p(\mathbf{x}_i)$ and $q_j=q(\mathbf{x}_j)$. $\mathbf{W}$ is a pairwise similarity matrix derived by labels, and $\varPi(\cdot)$ is a metric function which measures the similarities or distances of a pair of samples between Hamming space and original space. One of the alternatives for $\varPi$ is the similarity-similarity difference (means the difference between data similarity computed from Hamming space\footnote{In the training process, Hamming similarity between two samples is usually approximated by the product of the real-valued network output of one sample and the binary code of the other sample.} and data similarity derived from input space \cite{48}) function used in some deep asymmetric hashing methods \cite{7, 20, 38}:
\begin{align}
&\varPi\left(\ (p_i,q_j)\  | \ \mathbf{W}_{ij}\ \right)=||p(\mathbf{x}_i)^Tq(\mathbf{x}_j)-c\mathbf{W}_{ij}||^2 \label{label1}
\end{align}
In these deep asymmetric hashing methods, $p(\mathbf{x}_i)$ represents the output of a multi-layer neural network for data $\mathbf{x}_i$, $q(\mathbf{x}_j)$ represents the binary or multi-integer-valued code for data $\mathbf{x}_j$, $\mathbf{W}_{ij}=1$ if $\mathbf{x}_i$ and $\mathbf{x}_j$ are similar, $\mathbf{W}_{ij}=-1$, otherwise. The similarity-similarity difference function (\ref{label1}), however, may not well preserve the original similarity of data. Note that, the value of network output $p(\mathbf{x}_i)$ is continuous and thus it is not appropriate to minimize the gap between real-valued similarity $p(\mathbf{x}_i)^Tq(\mathbf{x}_j)$ and binary similarity $\mathbf{W}_{ij}$. To tackle this problem, we propose a novel distance-similarity product (means the product of data distance calculated from Hamming space and data similarity derived from input space) function for realizing $\varPi$ as follows:
\begin{align}
&\widetilde{\varPi}\left(\ (\widetilde{p}_i,\widetilde{q}_j)\  | \ \widetilde{\mathbf{W}}_{ij}\ \right)=||\widetilde{p}(\mathbf{x}_i)-\widetilde{q}(\mathbf{x}_j)||^2\widetilde{\mathbf{W}}_{ij}   \label{ASBL}
\end{align}
where $\widetilde{p}(\cdot)$ and $\widetilde{q}(\cdot)$ can be any suitable mapping functions, linear or nonlinear. The similarity matrix $\widetilde{\mathbf{W}}$ is constructed by an affinity graph $Gra\{\mathbf{X}; \widetilde{\mathbf{W}}\}$ that incorporates semantic knowledge of data, where $\widetilde{\mathbf{W}}_{ij}=1$ if $\mathbf{x}_i$ and $\mathbf{x}_j$ belong to the same class, $\widetilde{\mathbf{W}}_{ij}=0$, otherwise. To take the full advantage of feature learning ability, we adopt two neural networks parameterized by $\theta_1$ and $\theta_2$ as the nonlinear functions for $\widetilde{p}(\cdot)$ and $\widetilde{q}(\cdot)$. As a result, we obtain the following pairwise similarity preserving loss $\mathcal{P}$:
\begin{equation}
\min_{\theta_1,\theta_2}\mathcal{P}=\sum_{i}\sum_{j}\widetilde{\varPi}\left(\ (\widetilde{p}_i,\widetilde{q}_j)\  | \ \widetilde{\mathbf{W}}_{ij}\ \right)=\sum_{i}\sum_{j} ||\phi(\mathbf{x}_i;\theta_1)-\phi(\mathbf{x}_j;\theta_2)||^2\widetilde{\mathbf{W}}_{ij} \label{preserving}
\end{equation}
where $\widetilde{p}_i=\phi(\mathbf{x}_i;\theta_1)$ and $\widetilde{q}_i=\phi(\mathbf{x}_j;\theta_2)$ are the outputs of two deep neural networks. Obviously, the proposed pairwise similarity preserving $\mathcal{P}$ aims to minimize the distances between class-related real-valued embeddings based on the similarity matrix of affinity graph $Gra\{\mathbf{X}; \widetilde{\mathbf{W}}\}$.

\subsection{Class Structure Quantization} 
Given the real-valued embedding $f(\mathbf{x}_i)$ and binary code $\mathbf{h}_i$ of the $i$-th sample $\mathbf{x}_i$, existing deep hashing methods consider the quantization error between $f(\mathbf{x}_i)$ and $\mathbf{h}_i$ for each $i \in [1,n]$ by minimizing the following objective:
\begin{equation}
\min_{\mathbf{H}, f} \sum_{i}||\mathbf{h}_i-f(\mathbf{x}_i)||^2
\end{equation}
Apparently, such one-to-one quantization strategy performs class-unrelated quantization and the class prior of data is completely ignored in the quantization process.
In this paper, we argue that exploiting class structure information of data for supervised quantization is necessary for discriminative binary code learning. As illustrated in Figure \ref{class_structure_quantization}, it is desirable that the quantization errors between binary code $\mathbf{h}_i$ and all real-valued embeddings $f(\mathbf{x}_j)$ (which share the same class with $\mathbf{x}_i$) should be minimized, where $i \in [1,n]$, $j \in \kappa(i)$, and $\kappa(i)$ denotes the index set of the samples that have the same label with $\mathbf{x}_i$. To achieve this goal, we design the following quantization loss for discriminative binary code learning:
\begin{equation}
\min_{\mathbf{G}}\sum_{i}\sum_{j \in \kappa(i)}\mathbf{\mathbf{G}}_{ij}\mathbf{S}_{ij} \label{quantization1}
\end{equation} 
where $\mathbf{G}_{ij}=||\mathbf{h}_i - f(\mathbf{x}_j)||^2$, $\mathbf{S}_{ij}\in \{0,1\}$ is an indicator, if $\mathbf{x}_i$ and $\mathbf{x}_j$ belong to the same class, $\mathbf{S}_{ij}=1$, otherwise, $\mathbf{S}_{ij}=0$. Obviously, $\sum_j \mathbf{S}_{ij}=|\kappa(i)|$, where $|\kappa(i)|$ is the cardinality of index set $\kappa(i)$.
Considering the class imbalance problem, we add a factor on the quantization loss (\ref{quantization1}) to adjust the penalty weights of different classes:
\begin{equation}
\min_{\mathbf{G}}\sum_{i}\frac{1}{|\kappa(i)|}\sum_{j \in \kappa(i)}\mathbf{G}_{ij}\mathbf{S}_{ij} 
\end{equation} 
Note that, the indicator matrix $\mathbf{S}$ characterizes the class structure information of data, which is seamlessly integrated into the quantization process, tackling the class-unrelated quantization problem elegantly. By replacing the mapping function $f(\cdot)$ with the neural networks $\phi(\cdot;\theta)$ used in (\ref{preserving}), we obtain the following class structure quantization loss $\mathcal{Q}$:
\begin{equation}
\min_{\mathbf{H}, \theta_1,\theta_2}\mathcal{Q}=\sum_{i}\frac{1}{|\kappa(i)|}\sum_{j \in \kappa(i)}(\mathbf{\widehat{G}}_{ij}+\mathbf{\widetilde{G}}_{ij})\mathbf{S}_{ij}
\end{equation}
where $\mathbf{\widehat{G}}_{ij}=||\mathbf{h}_i - \mathtt{tanh}(\phi(\mathbf{x}_j;\theta_1))||^2$ and $\mathbf{\widetilde{G}}_{ij}=||\mathbf{h}_i - \mathtt{tanh}(\phi(\mathbf{x}_j;\theta_2))||^2$.
The $\mathtt{tanh}$ function is utilized to force the real-valued network outputs to be binary-like codes, thereby achieving more precise quantization.

\begin{algorithm}[!tb]
	\small
	\caption{Dual Semantic Asymmetric Hashing}
	\LinesNumbered
	\KwIn{Training set $\mathbf{X}$, dual semantic labels $\mathbf{Y}$ and $\mathbf{R}$, binary code length $c$, iterations $T_1, T_2$, training samples size $m$ in each iteration, parameters $\alpha_1$, $\alpha_2$, $\beta_1$, $\beta_2$;}
	\KwOut{Networks $\theta_1$ and $\theta_2$, binary code $\mathbf{H}$, regression matrices $\mathbf{M}_1$ and $\mathbf{M}_2$;}	
	Initialize $\mathbf{H}$ randomly, set $\mathbf{U}$ and $\mathbf{V}$ as zero matrices\;	
	Calculate $(\mathbf{Y}^T\mathbf{Y})^{-1}\mathbf{Y}^T$ and $(\mathbf{R}^T\mathbf{R})^{-1}\mathbf{R}^T$\;	
	\For{$outT =1:T_1$}{
		Update $\mathbf{M}_1$ according to (\ref{M1})\;		
		Update $\mathbf{M}_2$ according to (\ref{M2})\;		
		Select $m$ training samples, construct $\mathbf{W}$ and $\mathbf{S}$\;		
		\For{$innerT =1:T_2$}{
			Calculate $u_i$ by forward propagation\;				
			Calculate the gradient of $\mathbf{U}$ according to (\ref{D1})\;				
			Update $\theta_1$ by back propagation\;						
			Calculate $v_j$ by forward propagation\;				
			Calculate the gradient of $\mathbf{V}$ according to (\ref{D2})\;					
			Update $\theta_2$ by back propagation;\
		}	
		Update $\mathbf{H}$ according to (\ref{H});\
	}
\end{algorithm}

The overall objective function (\ref{obj}) of the proposed DSAH can be optimized by a well-designed iterative algorithm, which can be seen in the next section. 

\subsection{Optimization}
For computational efficiency, we randomly select $m$ samples in each iteration for network training. As a result, the pairwise similarity matrix $\widetilde{\mathbf{W}}\in \Re^{m\times m}$ is constructed by $m$ samples. In order to utilize the entire class information of data, we train the neural networks with at least $n/m$ iterations, where each iteration uses a completely different set of $m$ samples. For details, we summarize the designed iterative algorithm for the proposed DSAH in Algorithm 1.

For convenience, we set $\mathbf{U}=[\mathbf{u}_1,...,\mathbf{u}_m]^T\in \Re^{m\times c}$, $\mathbf{V}=[\mathbf{v}_1,...,\mathbf{v}_m]^T\in \Re^{m\times c}$, where $\mathbf{u}_i=\phi(\mathbf{x}_i;\theta_1)$ and $\mathbf{v}_j=\phi(\mathbf{x}_j;\theta_2)$. 
The proposed iterative algorithm updates one variable at a time by fixing the other variables.

\textbf{Fix $\mathbf{V}$, $\mathbf{H}$ and update $\mathbf{U}$.} 
By dropping the terms not related to $\mathbf{U}$, we obtain the following optimization problem:
\begin{equation}
\min_{\mathbf{U}} \  \alpha_1 \mathtt{tr}(\mathbf{U}^T\mathbf{DU}-2\mathbf{U}^T\widetilde{\mathbf{W}}\mathbf{V})+\alpha_2 \mathtt{tr}(-2\mathbf{H}^T\mathbf{S}\mathtt{tanh}(\mathbf{U})+\mathtt{tanh}(\mathbf{U})^T\mathtt{tanh}(\mathbf{U}))
\end{equation}
where $\mathtt{tr(\cdot)}$ is the trace of a matrix and $\mathbf{D} \in \Re^{m\times m}$ is a diagonal matrix with $\mathbf{D}_{ii} = \sum_j \widetilde{\mathbf{W}}_{ij}$.  Taking the partial derivative with respect to $\mathbf{U}$, we obtain the gradient of $\mathbf{U}$ as
\begin{equation}
\frac{\partial \mathcal{J}}{\partial \mathbf{U}} = \alpha_1(\mathbf{DU}-\widetilde{\mathbf{W}}\mathbf{V})+\alpha_2(\mathtt{tanh}(\mathbf{U})-\mathbf{S}^T\mathbf{H})\odot (1-\mathtt{tanh}(\mathbf{U})^2) \label{D1}
\end{equation}
where $\odot$ indicates the Hadamard product. Based on the gradient $\frac{\partial \mathcal{J}}{\partial \mathbf{U}}$, we can update the network parameters $\theta_1$ by back-propagation with chain rule.

\textbf{Fix $\mathbf{U}$, $\mathbf{H}$ and update $\mathbf{V}$.} 
Using the same strategy as that for updating $\mathbf{U}$, we calculate the gradient of $\mathbf{V}$ by
\begin{equation}
\frac{\partial \mathcal{J}}{\partial \mathbf{V}} = \alpha_1(\mathbf{DV}-\widetilde{\mathbf{W}}\mathbf{U})+\alpha_2(\mathtt{tanh}(\mathbf{V})-\mathbf{S}^T\mathbf{H})\odot (1-\mathtt{tanh}(\mathbf{V})^2) \label{D2}
\end{equation}
Similarly, the network parameters $\theta_2$ can be updated by back-propagation with chain rule.

\textbf{Fix $\mathbf{H}$ and update $\mathbf{M}_1$, $\mathbf{M}_2$.} Discard irrelevant items and we obtain the following minimization problem: 
\begin{equation}
\min_{\mathbf{M}_1,\mathbf{M}_2} \ \mathtt{tr}(\mathbf{M}_1^T\mathbf{Y}^T\mathbf{YM}_1-2\sqrt{\beta_1}\mathbf{M}_1^T\mathbf{Y}^T\mathbf{H})-\mathtt{tr}(\mathbf{M}_2^T\mathbf{R}^T\mathbf{RM}_2-2\sqrt{\beta_2}\mathbf{M}_2^T\mathbf{R}^T\mathbf{H})
\end{equation}
Taking the derivatives with respect to $\mathbf{M}_1$ and $\mathbf{M}_2$ and setting them equal to zeros, the solutions of $\mathbf{M}_1$ and $\mathbf{M}_2$ are
\begin{gather}
\mathbf{M}_1=\sqrt{\beta_1}(\mathbf{Y}^T\mathbf{Y})^{-1}\mathbf{Y}^T\mathbf{H} \label{M1}\\
\mathbf{M}_2=\sqrt{\beta_2}(\mathbf{R}^T\mathbf{R})^{-1}\mathbf{R}^T\mathbf{H} \label{M2}
\end{gather}

\textbf{Fix $\mathbf{U}$, $\mathbf{V}$, $\mathbf{M}_1$, $\mathbf{M}_2$ and update $\mathbf{H}$.} The subproblem for $\mathbf{H}$ is a maximization problem with discrete constraint: 
\begin{equation}
\max_{\mathbf{H}} \  \mathbf{H}^T\mathbf{Q} \qquad\mathrm{s.t.}\  \mathbf{H} \in \{-1,1\}^{n \times c},\ \mathbf{H}^T\mathbf{1}=0
\end{equation}
where $\mathbf{Q} = \alpha_2(\mathbf{S}\mathtt{tanh}(\mathbf{U})+\mathbf{S}\mathtt{tanh}(\mathbf{V})) + \sqrt{\beta_1} \mathbf{YM}_1-\sqrt{\beta_2} \mathbf{R}\mathbf{M}_2$. This problem can be solved directly by sorting the elements of $\mathbf{Q}$ by column in descending order:
\begin{equation}
\mathbf{H}_{ij}=\left\{
\begin{array}{lr}
1, \qquad  \quad \mathrm{if}\ \  j\in \tau(i)& \\
-1,\qquad  \ \mathrm{otherwise}&
\end{array}
\right.  \label{H}
\end{equation}
where $\tau(i)$ indicates an index set of the first $n/2$ maximal elements of $\mathbf{Q}_{:,j}$ and $\mathbf{Q}_{:,j}$ indicates the $j$-th column of $\mathbf{Q}$.

\subsection{Out-of-Sample Extension}
After the training process is completed, the binary codes of the training set can be obtained. For new test samples, we utilize the obtained networks to generate their binary codes. Specifically, for any new point $\widetilde{\mathbf{x}} \notin \mathbf{X}$, the corresponding  binary code $\widetilde{\mathbf{h}}$ can be generated by
\begin{equation}
\widetilde{\mathbf{h}} = \mathtt{sign}(\phi(\widetilde{\mathbf{x}};\theta))
\end{equation}
where $\theta$ can be $\theta_1$ or $\theta_2$, $\phi(\cdot)$ is the output of either of the two neural networks, and $\mathtt{sign(\cdot)}$ is the binary function.

\subsection{Weight Sharing and Training Acceleration}
For the proposed DSAH, if we adopt two distinguish deep neural networks with different weights as the asymmetric hashing functions, the training process will be time-consuming.
Thanks to weight sharing mechanism, we adopt the same weights for these two networks to accelerate the training process. For comparison, we call the proposed method using different weights as DSAH-1 and refer to the proposed method adopting the weight sharing mechanism as DSAH-2.
Note that, the proposed DSAH-2 is still an asymmetric hashing methods since we can directly obtain the binary codes of training set during the training process while the binary codes of test set can be generated by network forward propagation followed with the $sign$ function. That is to say, the binary codes of training and test samples are generated in an asymmetric way.

\subsection{Connection and Comparison}
In this section, we discuss the connections and differences between our approach and some related methods, including shallow hashing methods SDH, FSDH, deep symmetric hashing methods DPSH, DSDH, and deep asymmetric hashing methods DAPH, ADSH.	

In comparison with shallow methods: SDH and FSDH generally encode images by hand-crafted features, leading to the lack of feature learning ability. Moreover, they only use single label for semantic information integration, resulting in the ignorance of the inter-class separability of data. Instead, our approach exploits the feature learning ability of deep neural networks to learn effective features for hashing. To learn semantic-sensitive hash codes, the semantic knowledge of data is fully utilized in our approach by designing a dual semantic regression loss.

In comparison with deep symmetric methods: DPSH and DSDH perform class-unrelated quantization since they minimize the quantization error for each individual sample rather than the class-related samples. As a result, they may ignore the class prior of data during the quantization process, and thus cannot give discriminative feedback for network training. To mitigate this problem, our approach adopts a class structure quantization, which minimizes the quantization errors between each hash code and all its class-related real-valued embeddings.

In comparison with deep asymmetric methods: DAPH utilizes a negative log-likelihood loss to learn deep hashing functions. Instead, ADSH uses a similarity-similarity difference function, which minimizes a $L_2$ loss between the supervised similarity and inner product of query-database hash code pairs. Being different from DAPH and ADSH, our approach, based on an affinity graph that incorporates semantic knowledge of data, adopts a novel distance-similarity product function, which minimizes the product of data distance calculated from Hamming space and data similarity derived from input space.

\section{Experiments}
For comparison, seven traditional hashing methods including LSH \cite{8}, SH \cite{51}, BRE \cite{21}, ITQ \cite{10}, SP \cite{52}, SDH \cite{41}, FSDH \cite{13}, and six deep hashing methods including DPSH \cite{28}, DAPH \cite{40}, DDSH \cite{19}, ADSH \cite{20}, DAGH \cite{6}, and DSDH \cite{27} are selected in our experiments.

\subsection{Data sets}
We evaluate the retrieval performance of the proposed DSAH on the following three well-known data sets.

\textbf{CIFAR-10}\footnote{http://www.cs.toronto.edu/kriz/cifar.html} consists of 60,000 $32\times 32$ color images from 10 classes (each class contains 6,000 samples). We randomly select 59,000 images as the training set, and the rest 1,000 images are utilized for testing. 

\textbf{Fashion-MNIST}\footnote{https://github.com/zalandoresearch/fashion-mnist} contains 70,000 images from 10 classes. In our experiments, 6,000 images are randomly sampled from each class to form the training set, and the rest images are collected as the test set.

\textbf{NUS-WIDE}\footnote{http://lms.comp.nus.edu.sg/research/NUS-WIDE.htm} is a multi-label data set, where each image has at least one label. Following the experimental settings in \cite{28}, we select 21 most frequent concepts. The training set has over 190,000 images while test set contains 2,100 samples.

\begin{table*}
	\footnotesize
	\centering
	\setlength{\tabcolsep}{1.4mm}
	\caption{Performance comparison of different methods in MAP (\%) on all used data sets.}
	\begin{tabular}{l|cccc|cccc|cccc}
		\hline
		\hline
		\multirow{2}{*}{Method}&\multicolumn{4}{c|}{CIFAR-10}   &\multicolumn{4}{c|}{Fashion-MNIST} &\multicolumn{4}{c}{NUS-WIDE}   \\
		\cline{2-13}
		&12 &24 &32 &48  &12 &24 &32 &48  &12 &24 &32 &48   \\
		\hline
		LSH
		&13.90 &15.68 &14.40 &16.27 &25.94 &24.56 &27.35 &33.08 &45.72 &53.92 &56.71 &59.40 \\
		SH
		&20.27 &18.27 &17.94 &17.53 &35.56 &31.57 &31.84 &29.46 &56.79 &56.22 &56.63 &62.10 \\
		BRE
		&19.99   &20.77     &22.50    &25.31    &30.58   &40.53     &43.55    &42.42   &57.84  &60.15   &62.99   &65.62      \\
		ITQ
		&23.03  &19.62  &19.98    &20.93      &36.48  &36.39  &37.80    &39.83      &67.38  &69.58  &70.10      &73.15      \\
		SP
		&21.67 &21.71 &21.48 &23.42 &38.72 &42.60 &43.90 &44.57 &68.54 &71.89 &72.63 &73.86 \\
		SDH
		&54.70 &67.50 &68.34 &68.17 &64.40 &79.96 &80.43 &81.03 &77.69 &79.90 &79.94 &81.14 \\
		FSDH
		&61.52 &64.72 &67.17 &68.42 &78.58 &80.30 &81.09 &80.86 &73.86 &78.01 &78.58 &79.28 \\
		\hline
		DPSH
		&76.15 &79.65 &80.01 &81.26 &81.64 &82.78 &83.76 &84.98 &78.69 &81.93 &82.80 &83.32 \\
		DAPH
		&74.95 &78.80 &78.91 &77.31 &82.62  &83.92 &83.94 &82.58 &76.15 &79.80 &79.68 &80.64 \\
		DSDH
		&79.70 &82.20 &82.20 &84.44 &82.12 &84.42  &85.02 &85.20 &78.82	&81.82	&82.72	&83.74
		\\
		DDSH
		&76.29    &81.64    &82.21    &81.26    &82.72    &85.51    &85.48    &85.62        &75.45    &79.87    &80.54    &81.46 \\
		ADSH
		&87.52 &91.35 &92.68 &93.24 &90.34 &93.39 &93.93 &94.37 &84.61 &88.08 &89.69 &90.81 \\
		DAGH
		&93.09 &92.85 &93.26 &94.03 &94.06 &\textbf{94.74} &94.56 &94.80 &84.20	&87.86	&88.83	&90.63
		\\
		DSAH-1
		&\textbf{94.60} &\textbf{95.47} &\textbf{95.32} &\textbf{95.24} &\textbf{94.75} &\textbf{95.13} &\textbf{95.49} &\textbf{95.16} &\textbf{85.62} &\textbf{89.87} &\textbf{90.14} &\textbf{91.43} \\		
		DSAH-2
		&\textbf{93.99}   &\textbf{95.02}   &\textbf{95.14}   &\textbf{95.17}   
		&\textbf{94.41}   &{94.60}   &\textbf{95.32}   &\textbf{95.00}   
		&\textbf{87.60}   &\textbf{88.52}   &\textbf{89.86}   &\textbf{91.10}     \\	
		\hline
		\hline
	\end{tabular}
	\label{Performance_all}
\end{table*}

\subsection{Experimental Settings}
All traditional methods use deep features extracted by the pre-trained CNN-F network \cite{4} for training, while deep hashing methods utilize raw pixels as the inputs. Specifically, for traditional methods, we utilize PCA to reduce the 4096-dimensional deep features into 1000-dimensional feature vectors as inputs for computational efficiency. As a result, the input for traditional hashing methods is a $n\times 1000$-dimensional matrix, where $n$ represents the number of training samples. For deep hashing methods, all images are cropped to $224\times 224$ pixels and we directly use the raw image pixels as the inputs for network training. During the training process, traditional methods use the entire training samples to learn hashing functions.
Following the experimental settings in \cite{28}, deep hashing methods use 5,000 samples on CIFAR-10 and Fashion-MNIST, and 10,500 samples on NUS-WIDE to update networks. For all comparison methods, we re-run the corresponding codes released by the authors, and the parameter settings depend on the recommendation of the corresponding papers. 
Note that, there are no open sources for DAPH and DSDH, so we run these methods through our own implementations. 
For fairness, all deep hashing methods exploit the same CNN-F network pre-trained on ImageNet \cite{28}.

The proposed DSAH is implemented with deep learning toolbox MatConvNet \cite{46} on a GeForce GTX 1080 GPU server. In the experiments, we fix batch size to 64 and weight decay parameter to $5\times 10^{-4}$. The learning rate is tuned among $\{10^{-6}, \ldots, 10^{-2}\}$ and the number of bits is ranged from $\{12,24,32,48\}$ for all data sets. 
We empirically set $\alpha_1$, $\alpha_2$, $\beta_1$, $\beta_2$ to $10^{-2}$, $10^3$, $10^2$ and $10$, respectively. In our experiments, two images will be considered similar if they share at least one same label. 
We study two variants of the proposed DSAH. The DSAH-1 uses two networks with different weights for hashing function learning, while the DSAH-2 adopts weight sharing mechanism in the process of training.

For evaluation, we adopt Hamming ranking and hash lookup as the retrieval protocols. Four indicators are used to compare the retrieval performance of different methods, including Precision, Recall, Means Average Precision (MAP), and F-measure. The precision and recall results are calculated based on 2 Hamming distances. For NUS-WIDE, the MAP results are calculated based on top-5k returned neighbors as in \cite{19}. 
We also provide precision-recall curves to evaluate the retrieval performance of different methods.

\begin{table*}[!tb]
	\footnotesize
	\centering
	\setlength{\tabcolsep}{3.5mm}
	\caption{Network generalization performance of deep hashing methods in MAP (\%) and Precision (\%) on CIFAR-10 and Fashion-MNIST data sets.}
	\begin{tabular}{c|l|cccc|cccc}
		\hline
		\hline
		\multirow{2}{*}{}&\multirow{2}{*}{Method}&\multicolumn{4}{c|}{MAP}  &\multicolumn{4}{c}{Precision}  \\
		\cline{3-10}
		& &12 &24 &32  &48 &12 &24 &32 &48  \\
		\hline
		\multirow{7}*{\rotatebox[]{90}{CIFAR-10}}&DPSH 
		&76.15  &79.65  &80.01  &81.26   &79.21   &81.38   &78.13   &68.87      \\
		&DAPH  
		&74.95 &78.80 &78.91 &77.31   &66.07   &79.75   &80.03   &77.86   \\
		&DSDH  
		&79.70 &82.20 &82.20 &84.44    &79.89   &82.38   &80.74   &75.46     \\		
		&DDSH
		&76.29    &81.64    &82.21    &81.26   &78.94   &82.76   &82.01   &80.43    \\		
		&ADSH   
		&73.04    &85.27    &88.12     &90.00  
		&76.13    &84.93    &85.08     &84.26        \\
		&DAGH
		&87.73 &89.21 &89.94 &90.74  
		&87.95 &89.18 &89.19 &87.97 \\
		&DSAH-1   
		&\textbf{94.24}    &\textbf{95.26}   &\textbf{95.25}  &\textbf{95.16}
		&\textbf{92.51} &\textbf{92.83}  &\textbf{92.47} &\textbf{92.24}    \\
		&DSAH-2   
		&\textbf{93.77}    &\textbf{94.94}   &\textbf{95.02} &\textbf{95.10}  
		&\textbf{92.38}  &\textbf{92.61}  &\textbf{91.92} &\textbf{91.18}  \\
		\hline
		\multirow{7}*{\rotatebox[]{90}{Fashion-MNIST}} &DPSH 
		&81.64 &82.78 &83.76 &84.98 &83.80   &85.30   &84.63   &81.16   \\
		&DAPH
		&82.62  &83.92 &83.94 &82.58  &67.58   &82.20   &83.86   &80.18     \\
		&DSDH
		&82.12 &84.42  &85.02 &85.20   &82.69   &86.15   &85.86   &83.13    \\		
		& DDSH
		&82.72    &85.51    &85.48    &85.62   &83.94   &86.43   &86.51   &85.45     \\	
		&ADSH   	
		&80.80      &87.58     &89.31      &90.45  
		&81.10      &87.75     &88.34      &88.00        \\
		&DAGH
		&87.73 &89.21 &89.94 &90.74 
		&90.47 &91.59 &91.48 &90.42    \\
		&DSAH-1    
		&\textbf{91.86}     &\textbf{92.20}    &\textbf{92.90}   &\textbf{92.90} 
		&\textbf{91.83}     &\textbf{92.65}    &\textbf{92.73}   &\textbf{92.32}  \\
		&DSAH-2   
		&\textbf{91.70}     &\textbf{92.09}   &\textbf{93.12} & \textbf{92.68}   
		&\textbf{92.13} &\textbf{92.39}  &\textbf{92.75} &\textbf{92.33}    \\
		\hline
		\hline
	\end{tabular}
	\label{generalizationCIFAR}
\end{table*}

\subsection{Discussions}
The MAP results of different methods on the used three data sets are listed in Table \ref{Performance_all}. As we can see, supervised methods perform much better than unsupervised methods. For example, on CIFAR-10 data set with 12 bits code length, the MAP result of SDH is approximately 33\% higher than that of SP. Such performance improvement of supervised methods is benefiting from semantic information integration. In other words, compared with unsupervised hashing methods, supervised methods can learn semantic-sensitive hash codes by using semantic information of data to guide the learning of hashing function. In addition, deep hashing methods outperform traditional methods on the used three data sets in most cases. For instance, the MAP results of DPSH are at least 12 percentage points higher than that of FSDH on CIFAR-10 data set. The reason is that deep hashing methods can simultaneously perform feature learning and hashing function learning in an end-to-end framework, which helps to generate more effective feature for binary code learning.  
We also observe that deep asymmetric hashing methods, such as ADSH, DSAH-1 and DSAH-2, obtain higher MAP results than symmetric methods, such as DPSH, DSDH, DDSH, which indicates the superiority of asymmetric hashing mechanism. Moreover, the proposed DSAH-1 and DSAH-2 obtain best retrieval performance in terms of MAP indicator in most cases, which demonstrates that the binary codes learned by our methods are more  semantic-sensitive.

\begin{table*}[!tb]
	\footnotesize
	\centering
	\setlength{\tabcolsep}{3.5mm}
	\caption{Network generalization performance of deep hashing methods in Recall (\%) and F-measure (\%) on CIFAR-10 and Fashion-MNIST data sets.}
	\begin{tabular}{c|l|cccc|cccc}
		\hline
		\hline
		\multirow{2}{*}{}&\multirow{2}{*}{Method} &\multicolumn{4}{c|}{Recall} &\multicolumn{4}{c}{F-measure}  \\
		\cline{3-10}
		& &12 &24 &32 &48  &12 &24 &32 &48  \\
		\hline
		\multirow{7}*{\rotatebox[]{90}{CIFAR-10}}  &DPSH
		&67.40   &52.59   &46.39   &39.09   &72.83   &63.89   &58.21   &49.87   \\
		& DAPH  
		&80.84   &63.78   &55.03   &36.57   &72.71   &70.87   &65.22   &49.77   \\
		&DSDH  
		&73.28   &61.44   &58.36   &52.64   &76.44   &70.39   &67.75   &62.02   \\		
		&DDSH
		&63.73   &59.25   &61.16   &62.78   &70.52   &69.06   &70.06   &70.51   \\		
		&  ADSH  
		&58.66    &60.35    &60.37     &67.50   
		&66.26    &70.56    &70.62     &74.95       \\
		&DAGH
		&80.91 &79.02 &77.96 &76.99
		&84.28 &83.79 &83.20 &82.11 \\
		&DSAH-1     
		&\textbf{93.54} &\textbf{91.53}    & \textbf{91.95}  &\textbf{91.36}   
		& \textbf{93.02}  &\textbf{92.17}   &\textbf{92.21} & \textbf{91.80}  \\
		&DSAH-2   
		&\textbf{92.43} &\textbf{91.37}  &\textbf{91.68} &\textbf{90.60}   
		&\textbf{92.40} &\textbf{91.99}   &\textbf{91.80}  &\textbf{90.89}   \\
		\hline
		\multirow{7}*{\rotatebox[]{90}{Fashion-MNIST}} &DPSH  
		&77.08   &68.30   &66.29   &62.31   &80.30   &75.86   &74.35   &70.50   \\
		&DAPH
		&87.40   &85.29   &82.99   &84.13   &76.22   &83.71   &83.42   &82.11   \\
		&DSDH
		&81.25   &71.71   &65.96   &60.75   &81.96   &78.27   &74.61   &70.20   \\		
		&DDSH
		&76.76   &74.68   &75.93   &77.55   &80.19   &80.12   &80.88   &81.31   \\	
		&ADSH  
		&74.05      &72.63     &73.20      &75.24  
		&77.42      &79.48     &80.06      &81.12        \\
		&DAGH
		&86.86 &82.88 &83.11 &82.23
		&88.63 &87.02 &87.10 &86.13\\
		&DSAH-1  
		&\textbf{91.66}     &\textbf{90.63}    &\textbf{90.99}   &\textbf{90.77}   
		&\textbf{91.75}     &\textbf{91.63}    &\textbf{91.85}   &\textbf{91.54}  \\
		&DSAH-2    
		&\textbf{91.17} &\textbf{90.68}  &\textbf{90.87} &\textbf{90.89}   
		&\textbf{91.65} &\textbf{91.53}   &\textbf{91.80}  &\textbf{91.60}   \\
		\hline
		\hline
	\end{tabular}
	\label{generalizationCIFAR2}
\end{table*}

\subsection{Generalization of Network}
The deep hashing methods ADSH, DAGH, the proposed DSAH-1 and the proposed DSAH-2 can directly learn binary codes of the training set during the training process. To further evaluate the effectiveness of the proposed methods, we compare the network generalization of these methods. In this experimental setting, the above-mentioned methods generate binary codes for training set $X_{train}$ via $H_{train}=\mathtt{sign}(\phi(X_{train};\theta))$.
Tables \ref{generalizationCIFAR} and \ref{generalizationCIFAR2} give the MAP, precision, recall and F-measure results of all deep hashing methods on CIFAR-10 and Fashion-MNIST data sets. From Tables 3 and 4, we observe that the proposed DSAH-1 obtains the best network generalization and the proposed DSAH-2 achieves comparable performance with DSAH-1.
The potential reason is that integrating intra-class compactness and inter-class separability of data promotes our model to generate semantic-sensitive binary codes. Meanwhile, performing class structure quantization makes our model to give discriminative feedback for network training, thereby providing a significant improvement of generalization of network.

Comparing Table \ref{generalizationCIFAR} with Table \ref{Performance_all}, under the MAP criterion, the asymmetric retrieval performances of ADSH, DAGH, DSAH-1 and DSAH-2 are superior to their symmetric ones. 
For instance, on CIFAR-10 data set, when the code length is 12 bits, the MAP value of ADSH under the asymmetric setting is 87.52, which is about 14 percentage points higher than that under the symmetric setting (73.04). 
That is to say, using network forward propagation followed with $\mathtt{sign}$ function to obtain the binary codes of training set will result in significant performance degradation for ADSH. 
This is because using $\mathtt{sign}$ function to transform binary-liked real-valued network outputs into binary codes will make a large information loss. On the contrary, under the asymmetric settings, ADSH, DAGH, DSAH-1 and DSAH-2 can directly obtain the binary codes of training set during network training, which avoids such information loss caused by $\mathtt{sign}$ function.

\begin{figure*}[!tb]
	\centering
	\subfigure[\label{fig:CIFAR_12}]{\includegraphics[scale=0.19,trim=20 0 45 29,clip]{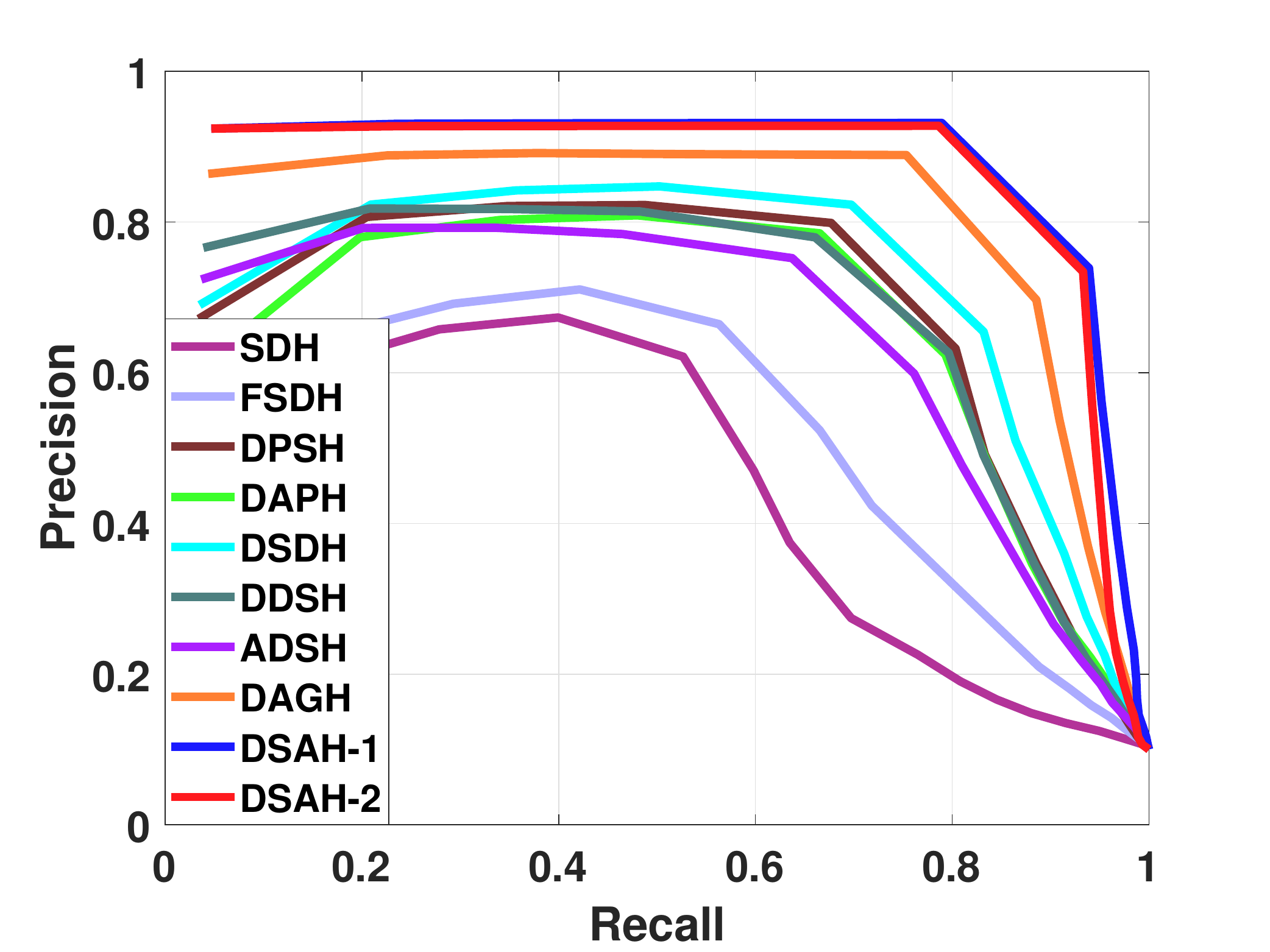}}
	\subfigure[\label{fig:CIFAR_24}]{\includegraphics[scale=0.19,trim=20 0 45 29,clip]{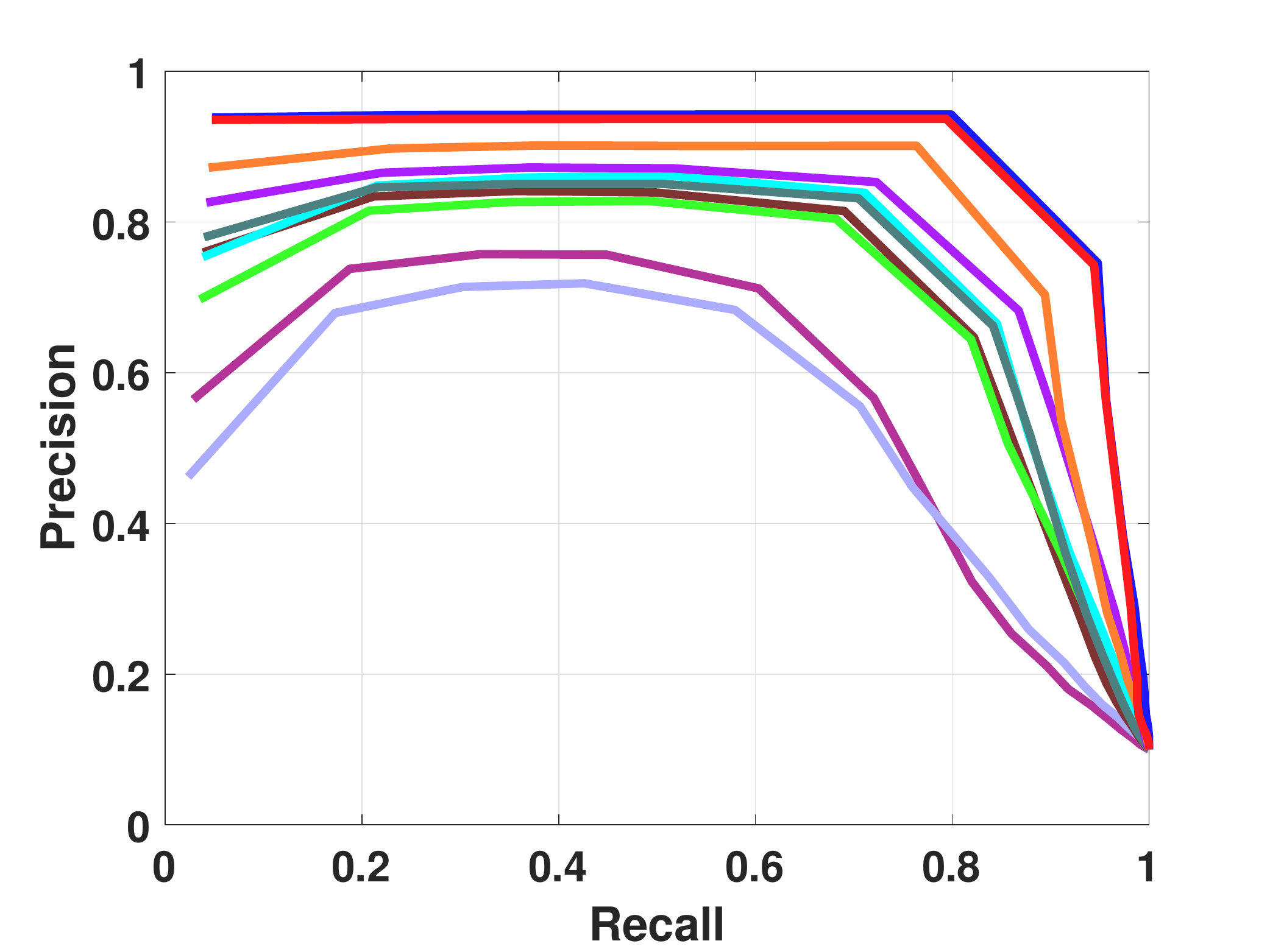}}
	\subfigure[\label{fig:CIFAR_32}]{\includegraphics[scale=0.19,trim=20 0 45 29,clip]{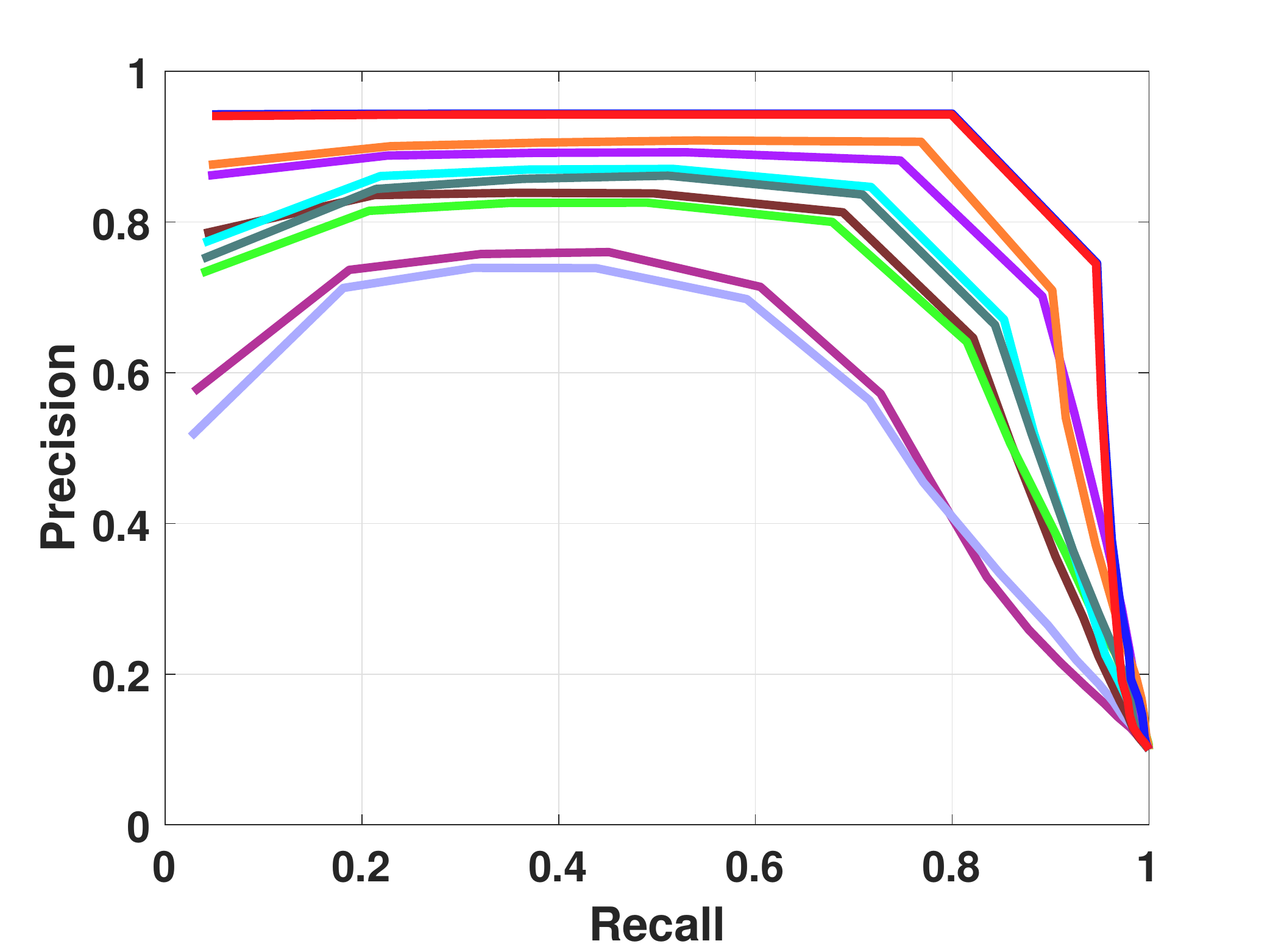}}18
	\subfigure[\label{fig:CIFAR_48}]{\includegraphics[scale=0.19,trim=20 0 45 29,clip]{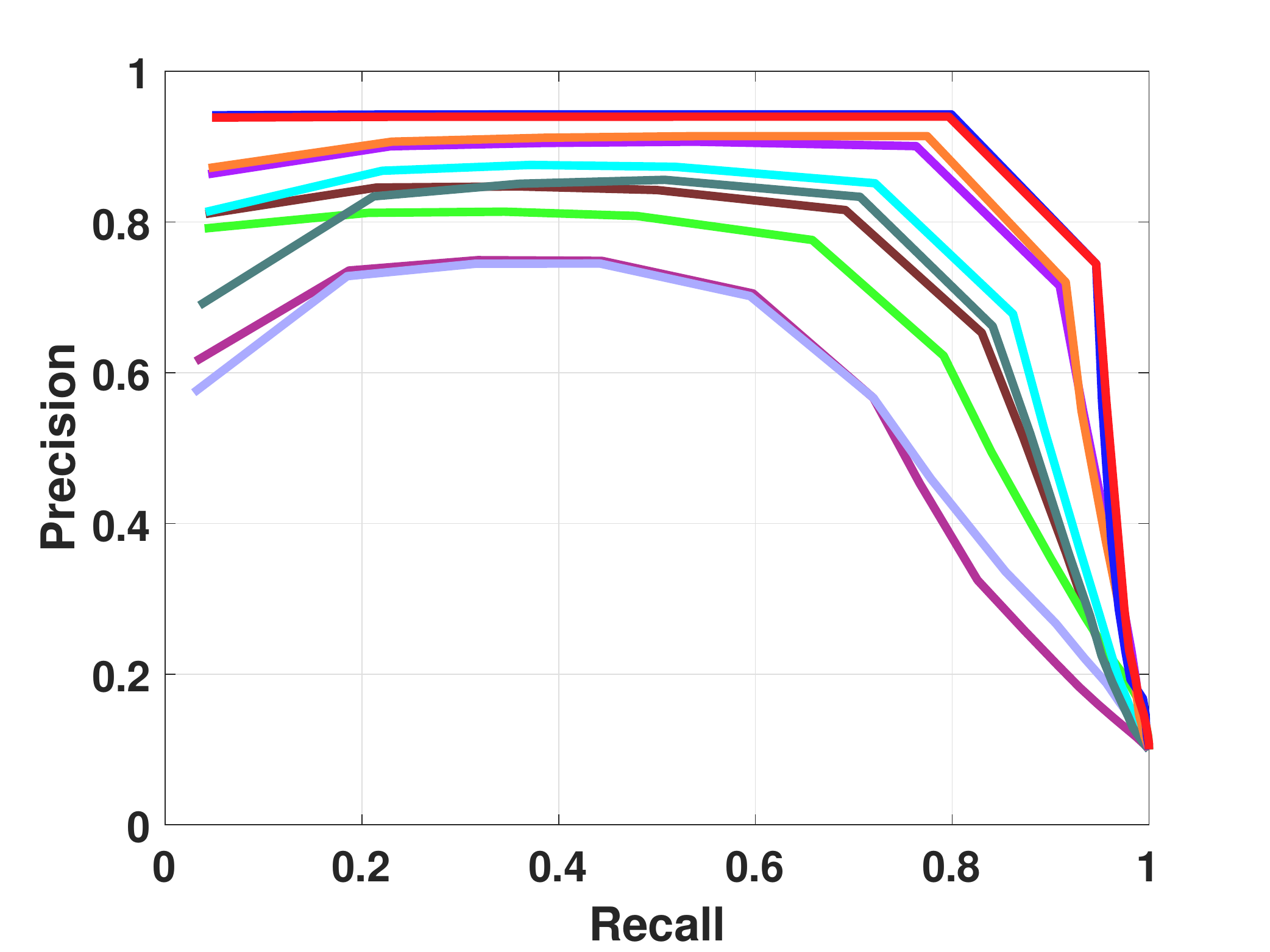}}
	\caption{Precision-recall curves of supervised hashing methods on CIFAR-10 data set with (a) 12, (b) 24, (c) 32, and (d) 48 bits.} 	\label{PRCIFAR}
\end{figure*}

\begin{figure*}[!tb]
	\centering
	\subfigure[\label{fig:PR_Fmnist_12}]{\includegraphics[scale=0.19,trim=20 0 45 29,clip]{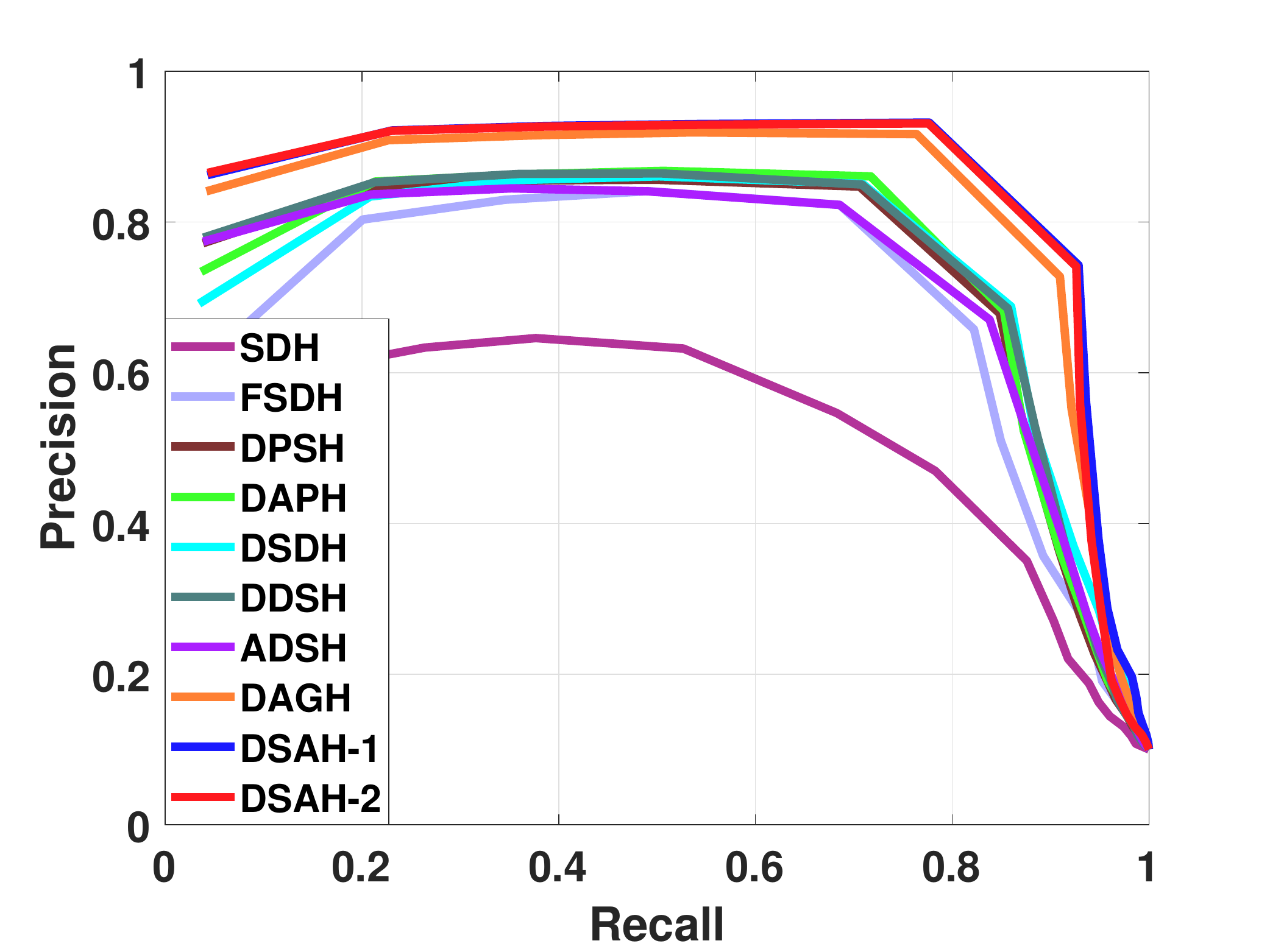}}
	\subfigure[\label{fig:PR_Fmnist_24}]{\includegraphics[scale=0.19,trim=20 0 45 29,clip]{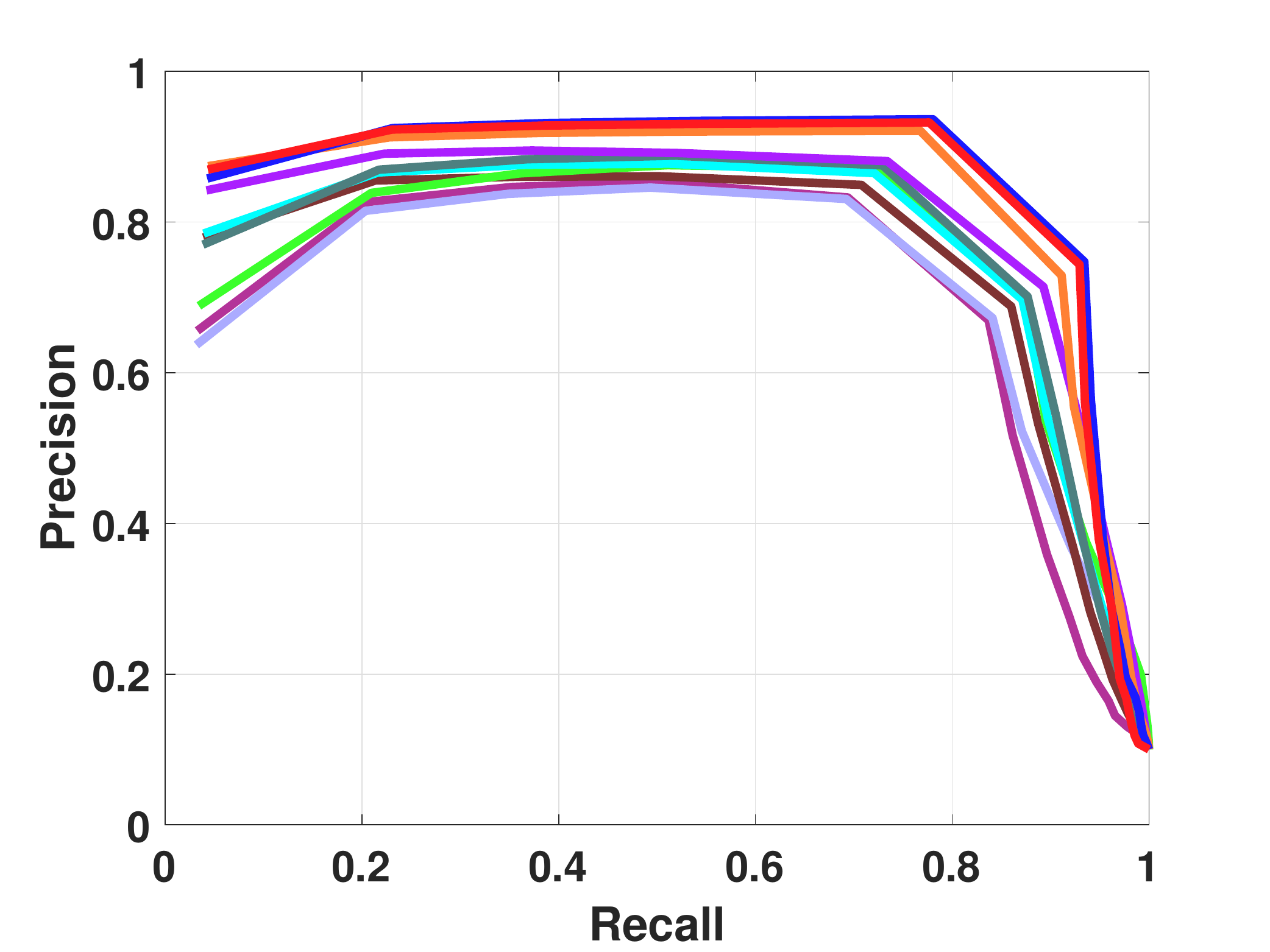}}
	\subfigure[\label{fig:PR_Fmnist_32}]{\includegraphics[scale=0.19,trim=20 0 45 29,clip]{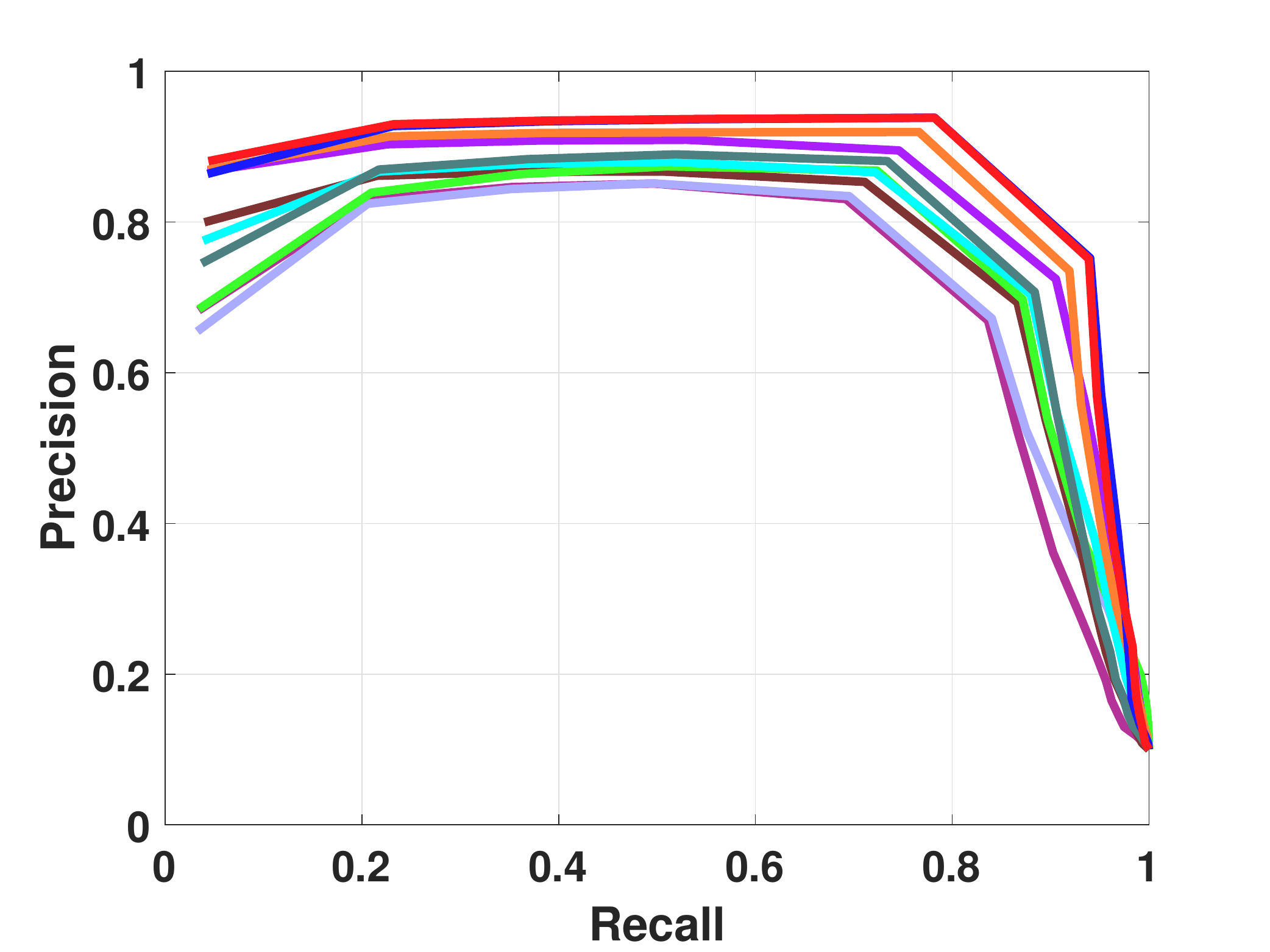}}
	\subfigure[\label{fig:PR_Fmnist_48}]{\includegraphics[scale=0.19,trim=20 0 45 29,clip]{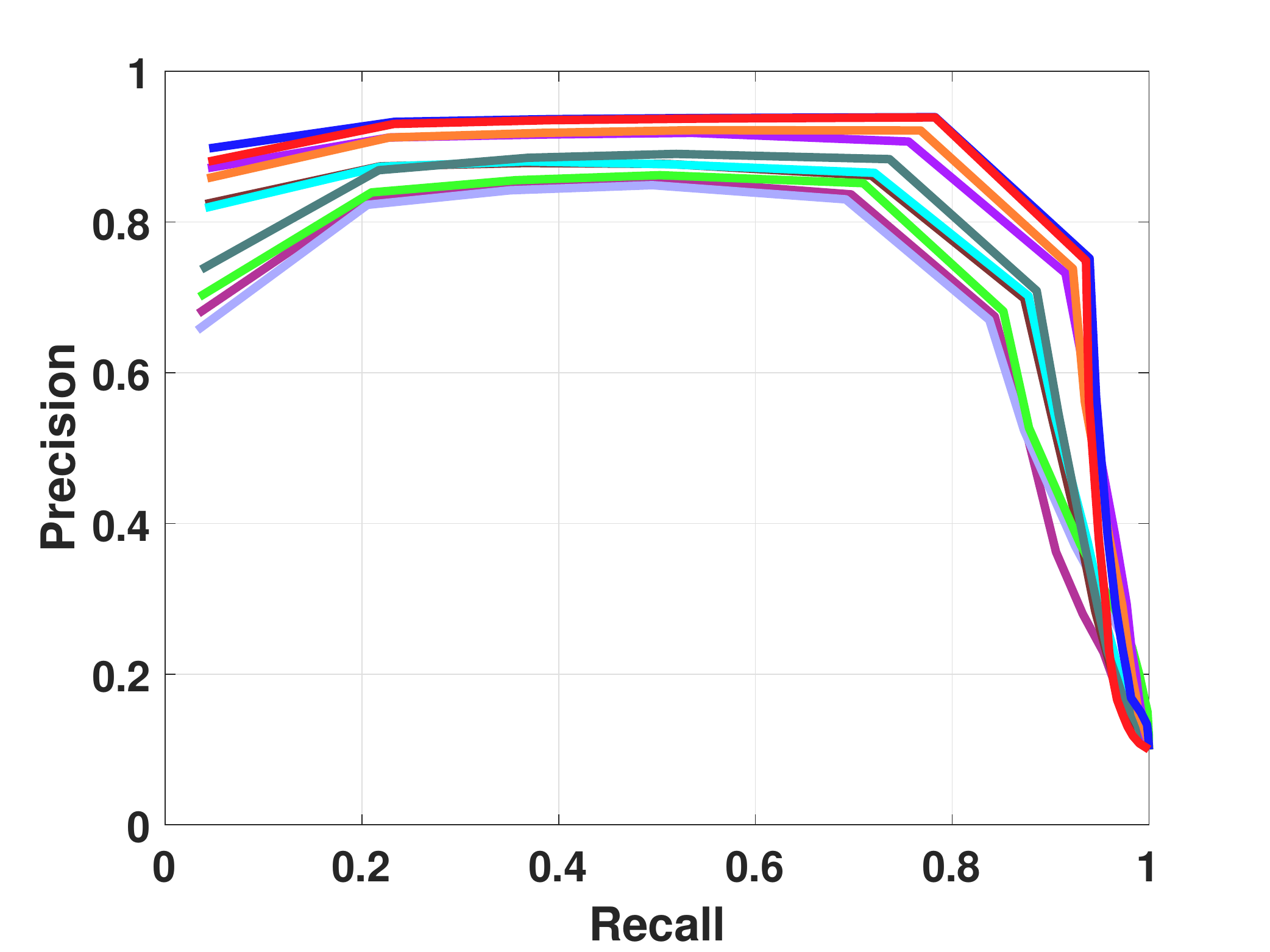}}
	\caption{Precision-recall curves of supervised hashing methods on Fashion-MNIST data set with (a) 12, (b) 24, (c) 32, and (d) 48 bits.} 	\label{PRFmnist}
\end{figure*}

We also show PR curves of supervised hashing methods on CIFAR-10 and Fashion-MNIST data sets, which can be seen in Figures \ref{PRCIFAR} and \ref{PRFmnist}, respectively. As we can see, when the length of hash code is very small, such as 12 bits, SDH may not achieve satisfactory performance, which can be also observed in Table \ref{Performance_all}. The potential reason is that 12 bits are too short for SDH to obtain satisfactory retrieval performance due to the DCC optimization algorithm, which requires SDH to obtain binary codes bit by bit. In addition, we observe that the proposed DSAH-1 and DSAH-2 obtain stable and promising performance with varying hash code lengths, which further demonstrates the effectiveness of our framework.


\subsection{Parameter Sensitivity}
\begin{figure*}[!tb]
	\centering
	\subfigure[\label{fig:alpha1}]{\includegraphics[scale=0.20,trim=330 35 360 55,clip]{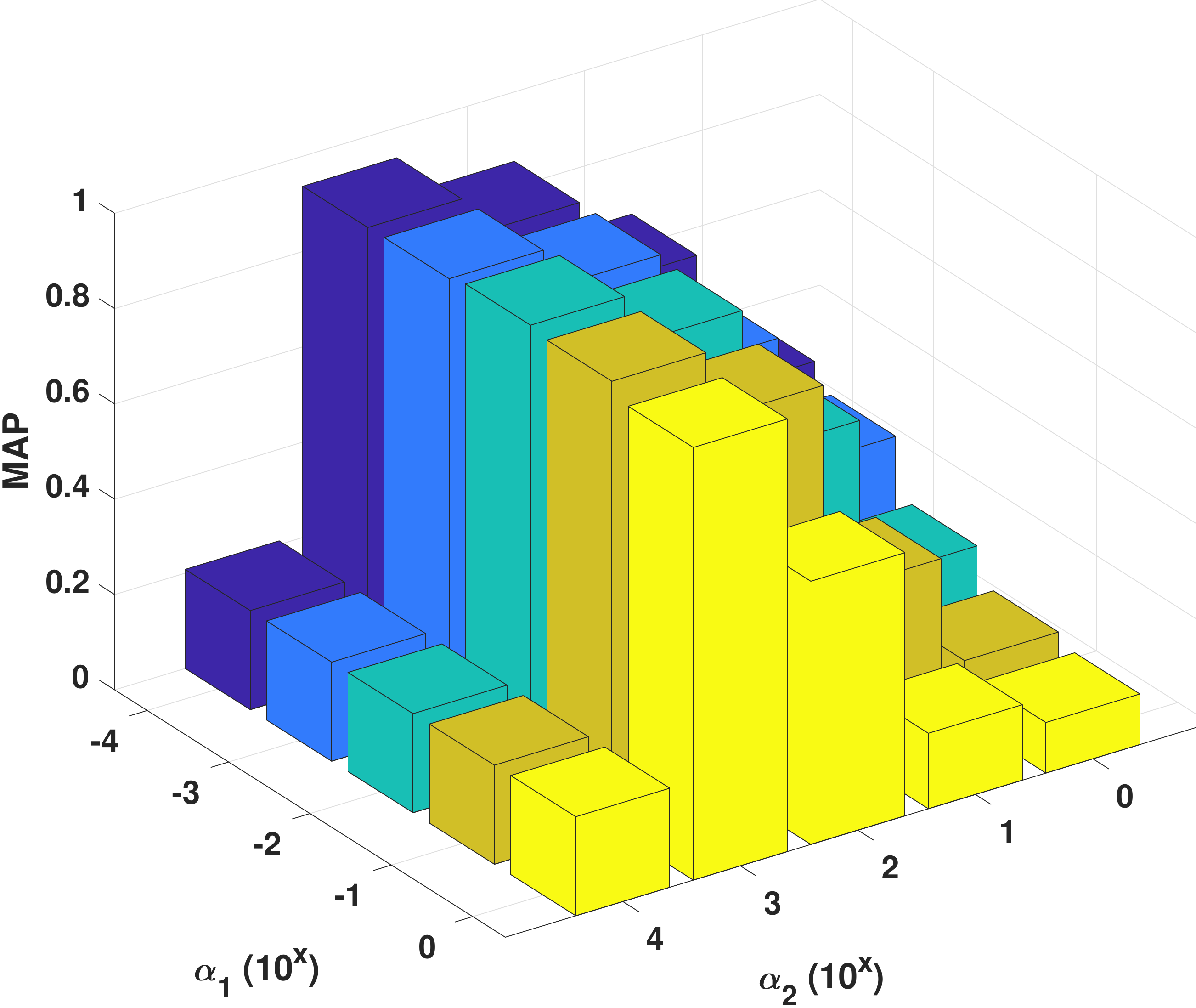}}
	\qquad\subfigure[\label{alpha1}]{\includegraphics[scale=0.20,trim=330 35 360 55,clip]{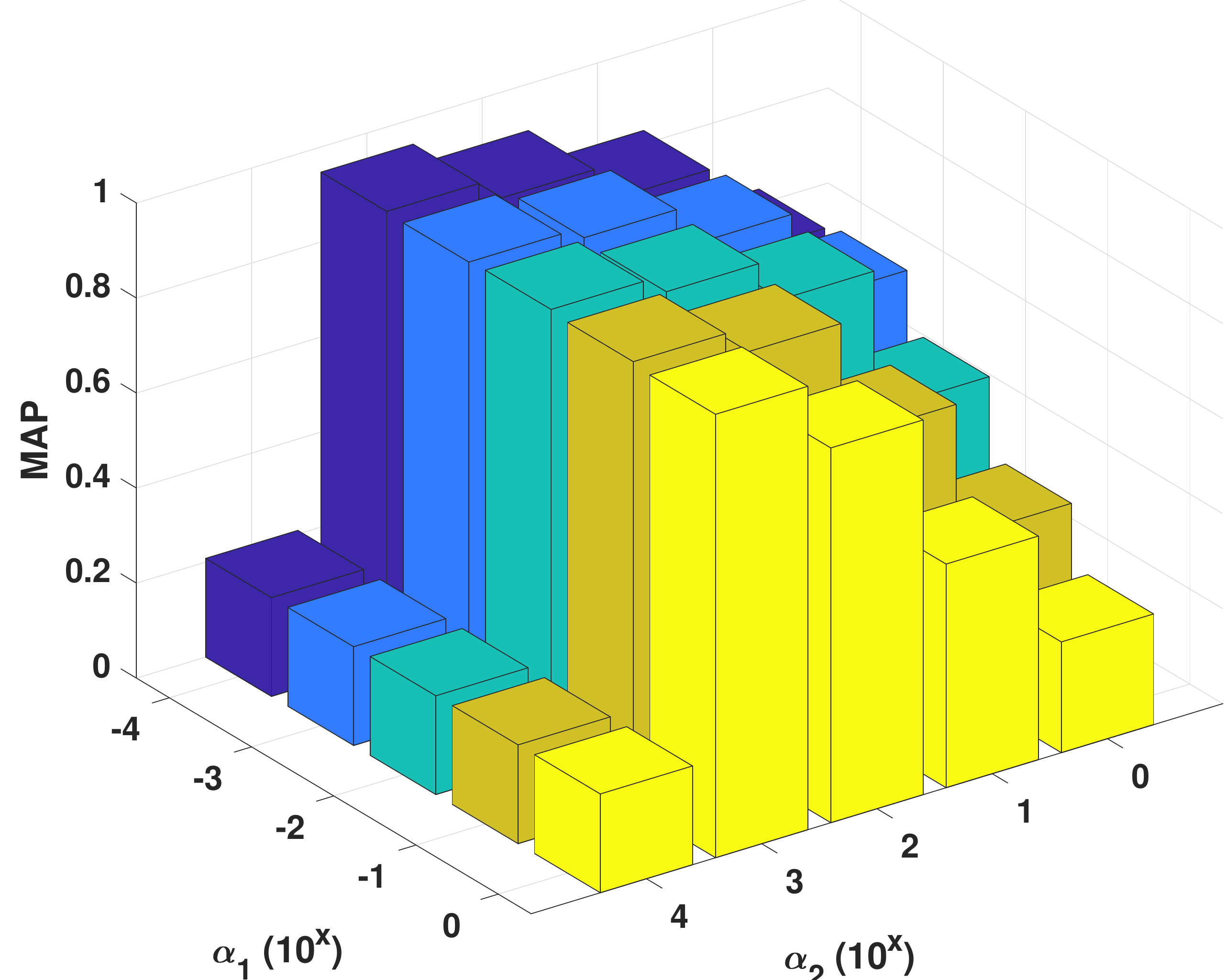}}
	\subfigure[\label{fig:alpha2}]{\includegraphics[scale=0.20,trim=328 35 360 55,clip]{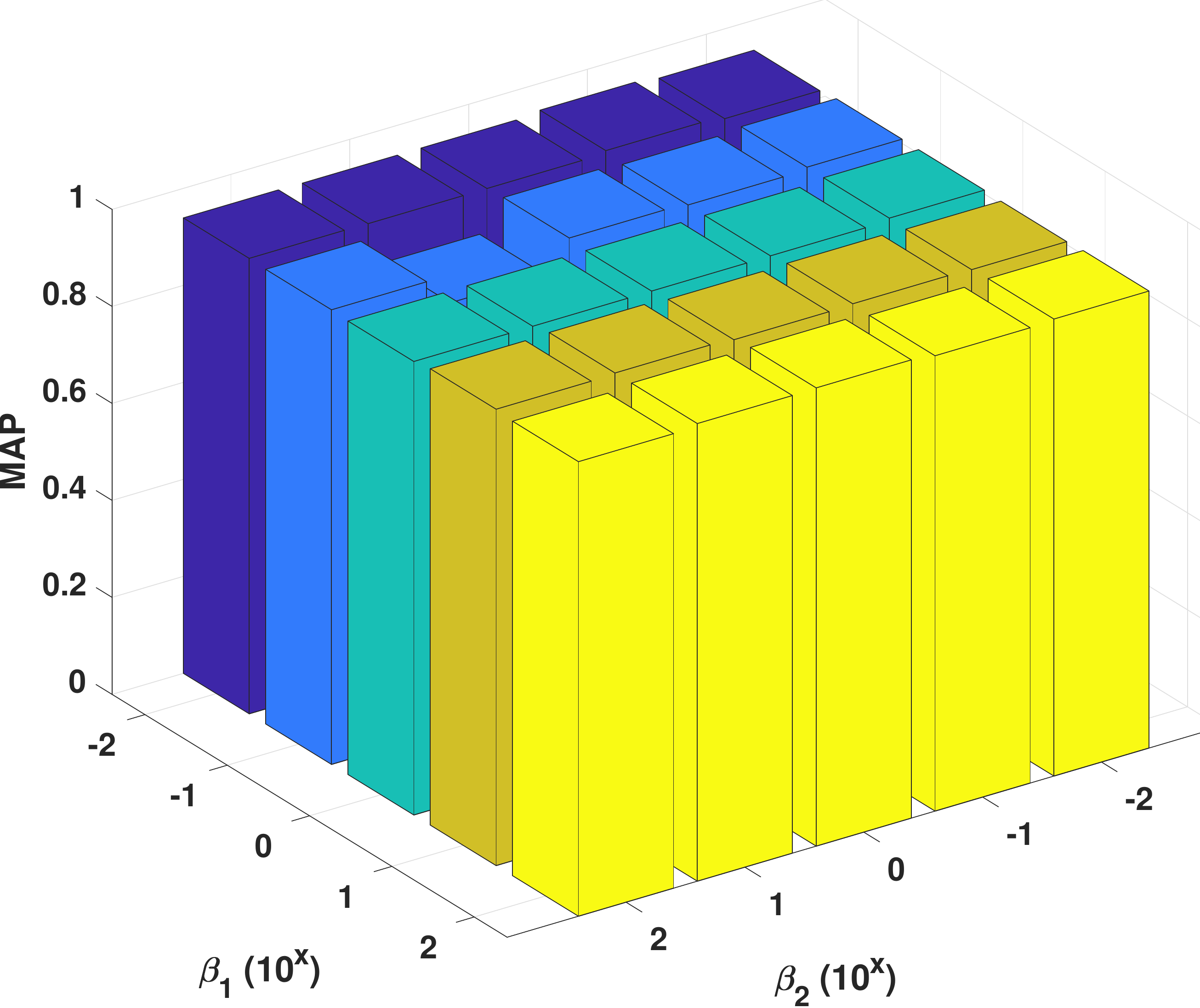}}
	\qquad\subfigure[\label{alpha2}]{\includegraphics[scale=0.20,trim=328 35 360 55,clip]{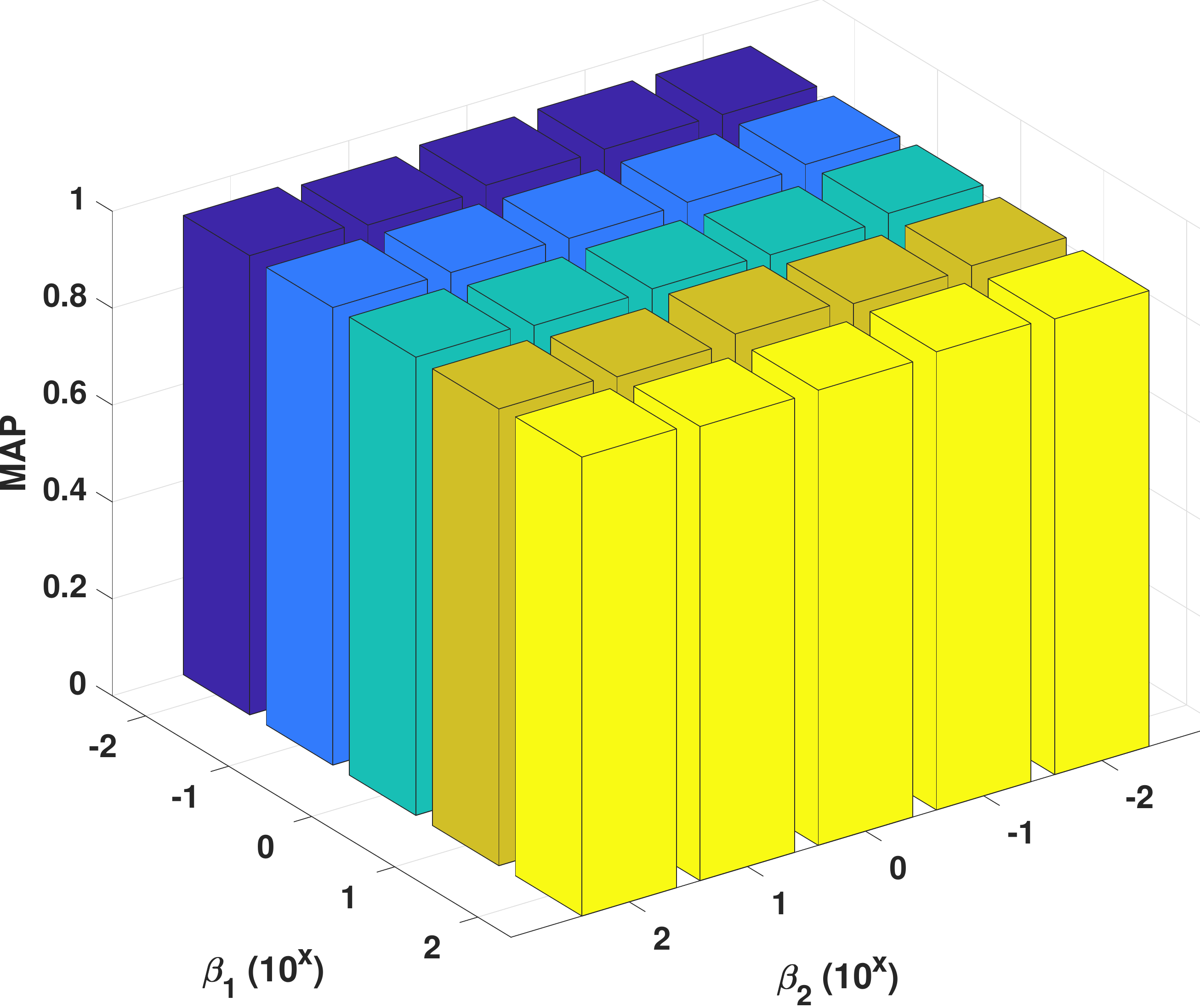}} 
	\caption{Parameter sensitivity. The values of MAPs versus $\alpha_1$ and $\alpha_2$ for the proposed (a) DSAH-1 and (b) DSAH-2. The values of MAPs versus $\beta_1$ and $\beta_1$ for the proposed (c) DSAH-1 and (d) DSAH-2.} 	\label{sensitivity}
\end{figure*}

Figure \ref{sensitivity} shows the parameter sensitivity of the proposed DSAH-1 and DSAH-2 on CIFAR-10 data set. The values of $\alpha_1$ and $\alpha_2$ are selected from the sets $\{10^{-4},\ldots,10^0\}$ and $\{10^{0},\ldots,10^4\}$, respectively. The values of parameters $\beta_1$ and $\beta_2$ are both tuned from the set $\{10^{-2},\ldots,10^2\}$. 
From Figure \ref{fig:alpha1} and (b), when $\alpha_2 \leq 10^3$, the performances of DSAH-1 and DSAH-2 improve with the increase of $\alpha_2$. However, when $\alpha_2 = 10^4$, the MAP values drop sharply. The potential reason is that, when $\alpha_2$ is very large, it will cause the gradient explosion problem. 
From Figure \ref{fig:alpha2} and (d), we observe that the proposed DSAH-1 and DSAH-2 are not sensitive to $\beta_1$ and $\beta_2$ in a large range with $10^{-2} < \beta_1, \beta_2 < 10^{2}$.

\begin{table*}[!tb]
	\footnotesize
	\centering
	\caption{Training time (in minutes) of deep hashing methods on all used data sets.}
	\label{time}
	\setlength{\tabcolsep}{1.6mm}
	\begin{tabular}{l|cccc|cccc|cccc}
		\hline
		\hline
		\multirow{2}{*}{Method}&\multicolumn{4}{c|}{CIFAR-10}     &\multicolumn{4}{c|}{Fashion-MNIST}   &\multicolumn{4}{c}{NUS-WIDE}   \\
		\cline{2-13}
		&12  &24  &32 &48 &12  &24  &32 &48 &12  &24  &32 &48 \\
		\hline
		DPSH
		&28.53     &28.22     &28.35     &28.76     &26.96     &27.10     &26.87     &27.39     &70.63   &74.02     &75.15     &77.78   \\
		DAPH
		&{68.85}     &{65.05}     &{65.64}   &{67.55}     &{55.94}     &{55.01}   &{55.28}     &{60.30}     &{167.23}  &{166.96}    &{165.16}    &{165.42}   \\
		DSDH
		&29.09     &29.97     &30.97     &34.33     &24.13     &24.12     &24.73     &26.72     &79.65   &80.93     &83.57     &84.17   \\
		DDSH
		&23.53     &24.10     &24.99     &26.10     &22.44     &22.98     &24.54     &25.34     &63.98   &67.04     &68.24     &72.84   \\
		ADSH
		&24.69     &30.95     &35.76     &47.60     &24.27     &31.59   &36.08     &47.44     &64.60     &85.01     &96.69     &128.48   \\ 
		DAGH
		&30.07     &30.11     &30.15     &30.64     &28.93     &29.20     &28.97     &29.42     &88.67     &88.92     &90.28     &90.92   \\ 
		DSAH-1
		&{62.58}     &{61.77}     &{63.00}   &{62.01}    &{54.72}     &{54.35}   &{53.78}     &{54.48}     &{162.97}  &{163.55}      &{162.22}  &{164.72}    \\
		DSAH-2
		&27.21     &27.75     &27.57     &27.90     &26.54     &26.77     &27.03     &27.02    &91.38     &90.09     &89.40      &89.82    \\
		\hline
		\hline
	\end{tabular}
\end{table*}

\subsection{Time Performance}
Table \ref{time} lists the training time of deep hashing methods on all used data sets. We observe that the training time of ADSH increases as the length of binary code increases. This is because ADSH adopts DCC optimizer which generates binary codes bit by bit. 
Besides, DAPH and DSAH-1 need more time for network training. The reason is that, compared with other deep hashing methods, DAPH and DSAH-1 need to train two deep neural networks as the asymmetric hashing functions.

\begin{figure*}[!tb]
	\centering
	\subfigure[\label{fig:DPSH}]{\includegraphics[scale=0.29,trim=36 20 34 20,clip]{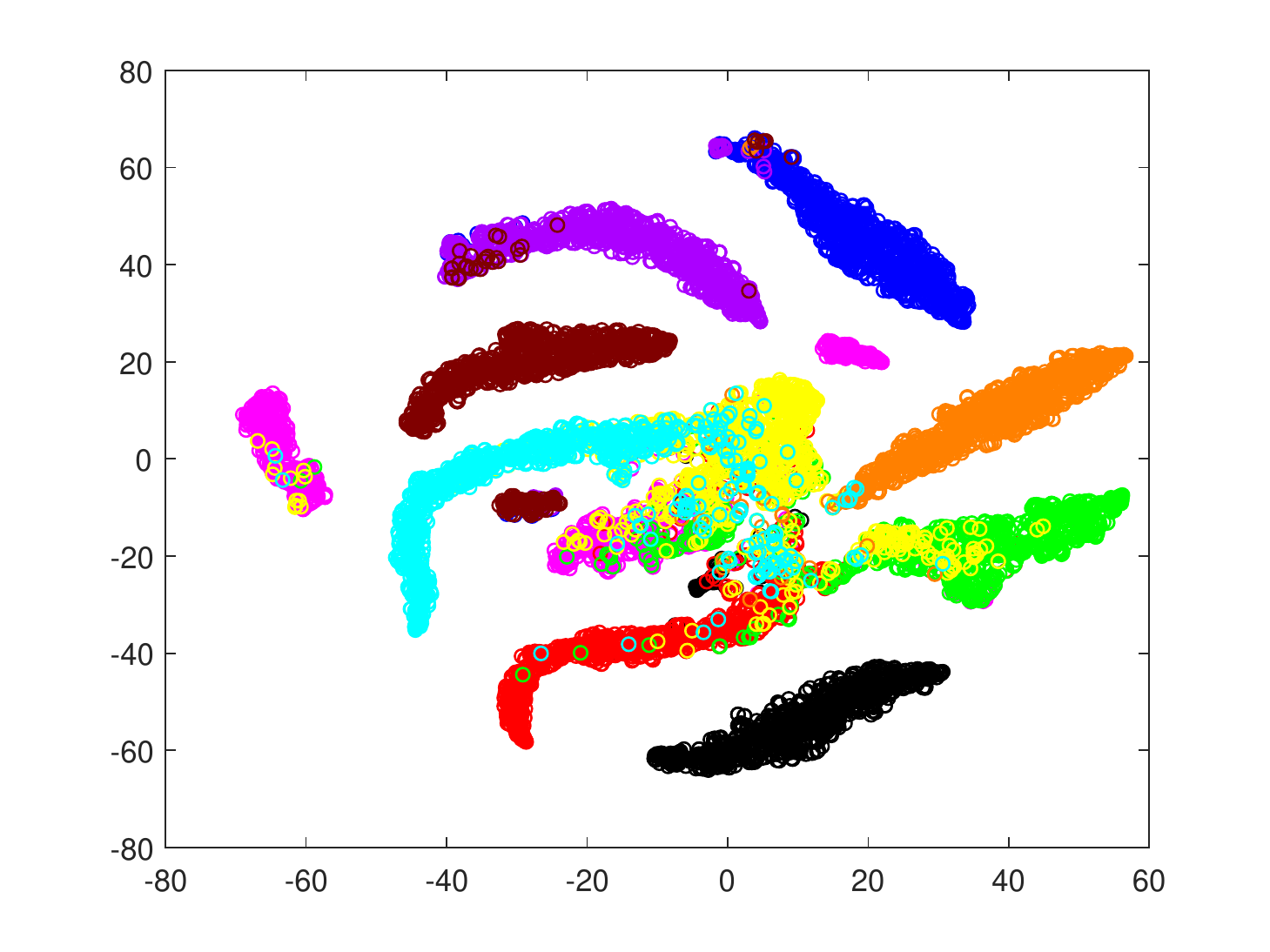}}	
	\subfigure[\label{fig:DAPH}]{\includegraphics[scale=0.29,trim=36 20 34 20,clip]{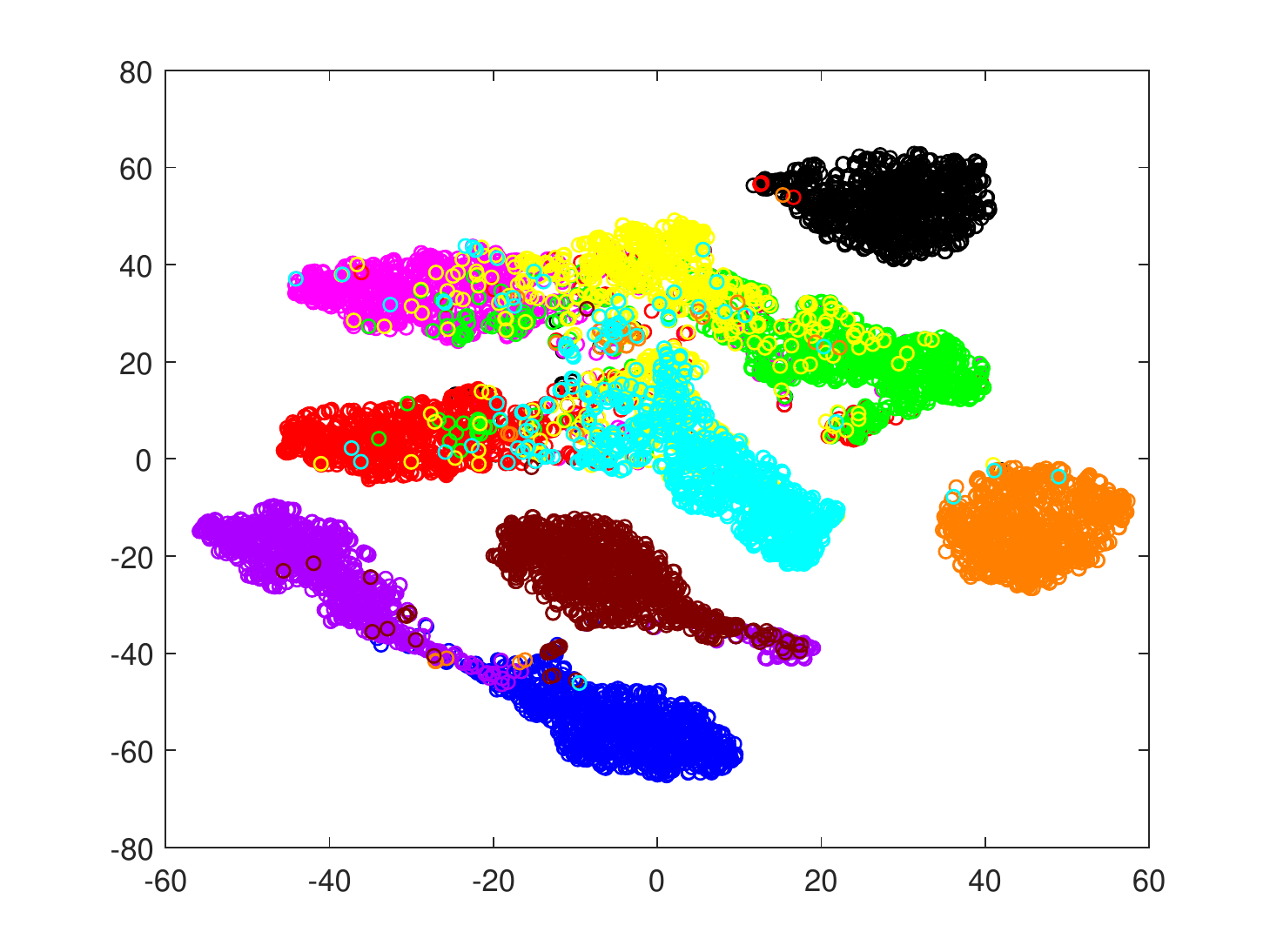}} 
	\subfigure[\label{fig:DSDH}]{\includegraphics[scale=0.29,trim=36 20 34 20,clip]{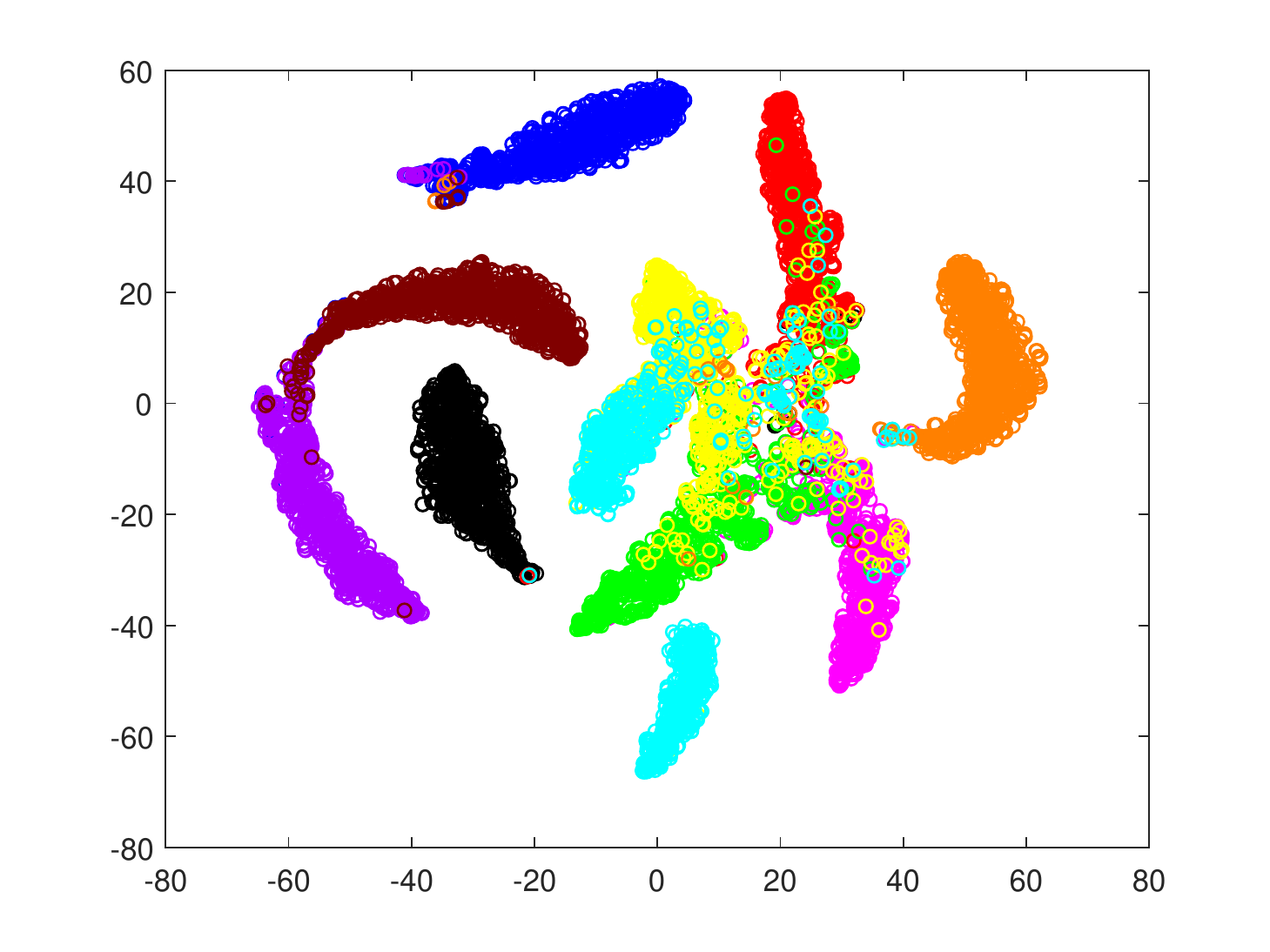}}
	\subfigure[\label{fig:DDSH}]{\includegraphics[scale=0.29,trim=36 20 34 20,clip]{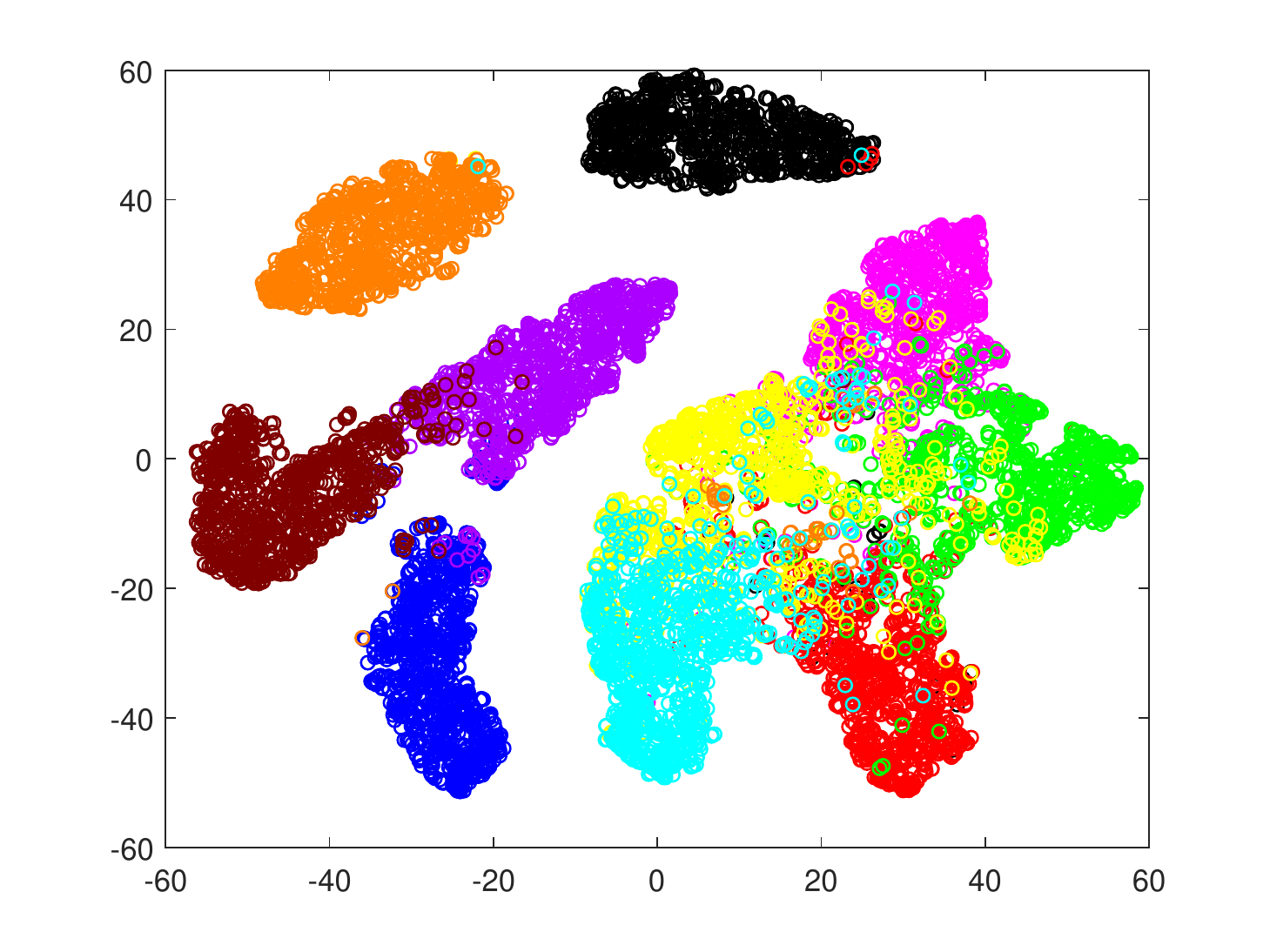}}
	\subfigure[\label{fig:ADSH}]{\includegraphics[scale=0.29,trim=36 20 34 20,clip]{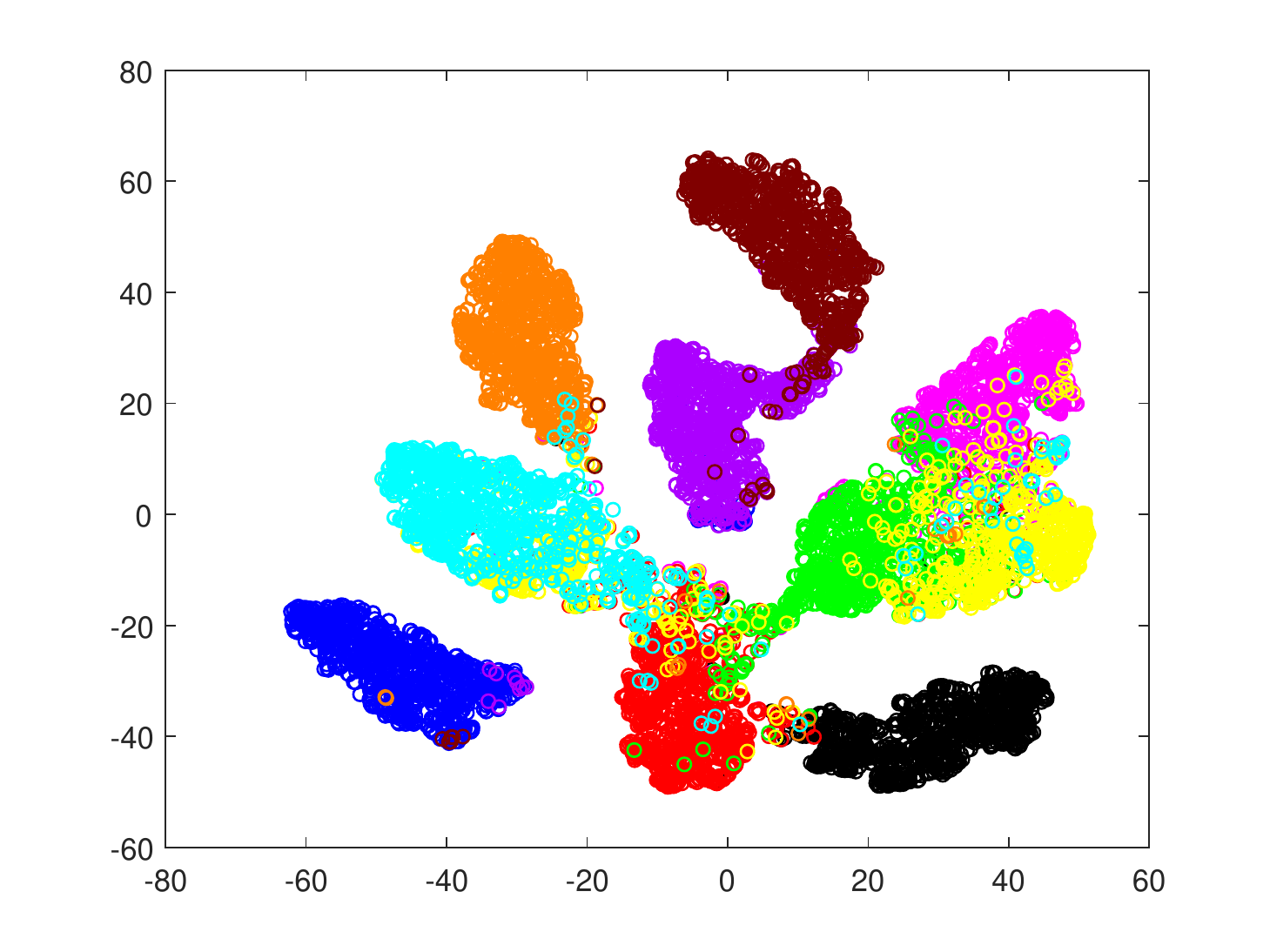}}
	\subfigure[\label{fig:DAGH}]{\includegraphics[scale=0.29,trim=36 20 34 20,clip]{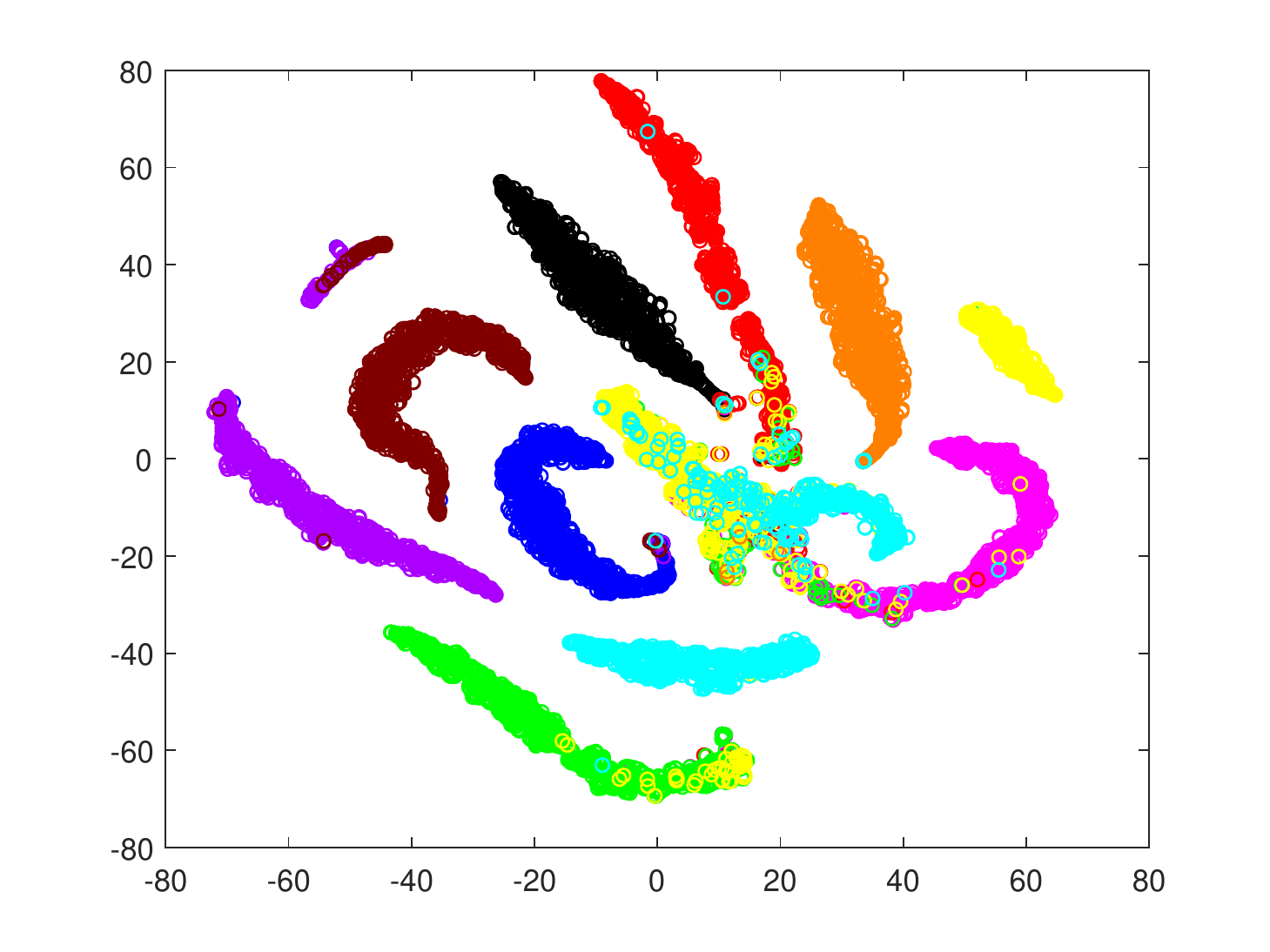}}
	\subfigure[\label{fig:DSAH-1}]{\includegraphics[scale=0.29,trim=36 20 34 20,clip]{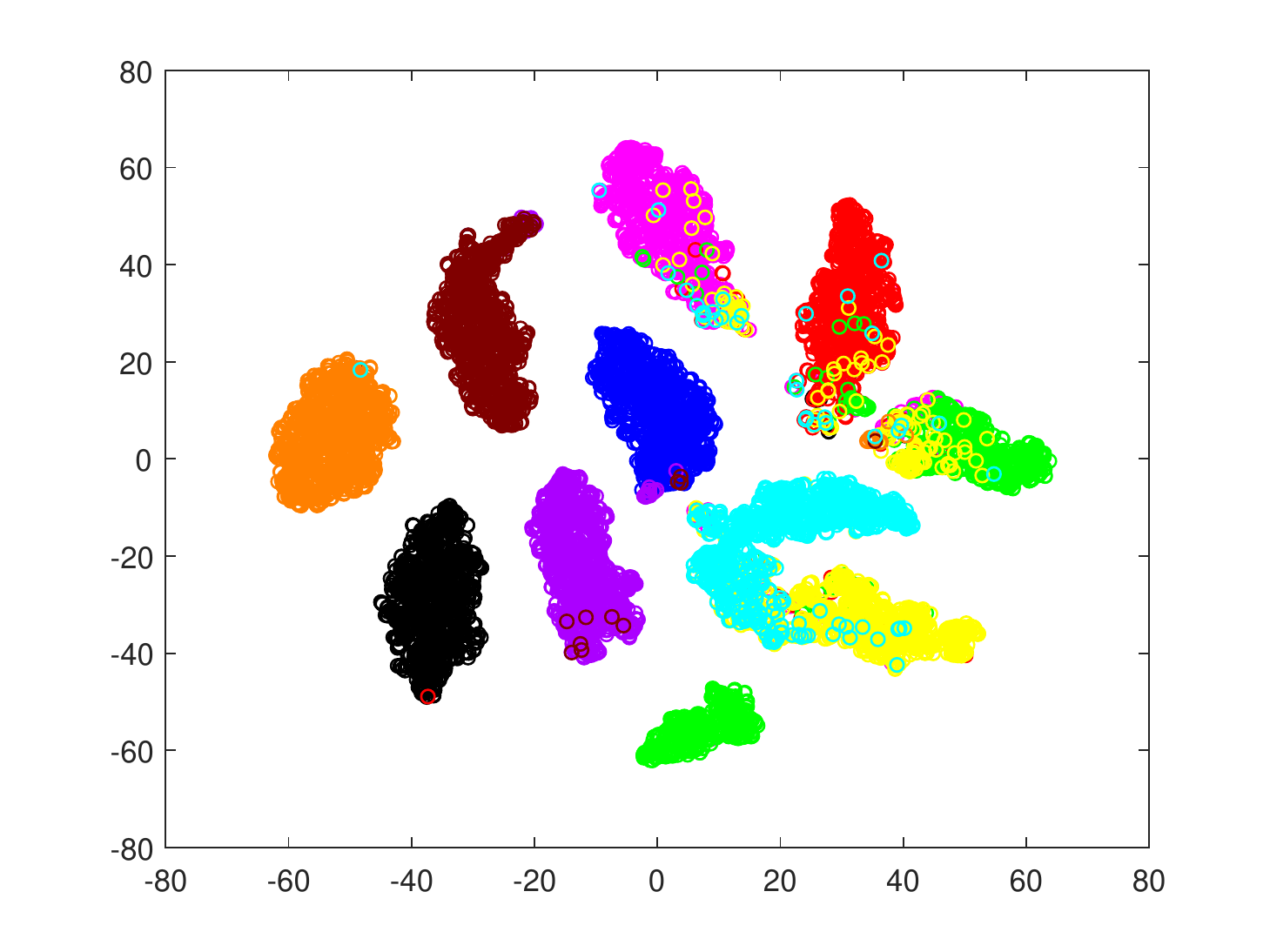}}
	\subfigure[\label{fig:DSAH-2}]{\includegraphics[scale=0.29,trim=36 20 34 20,clip]{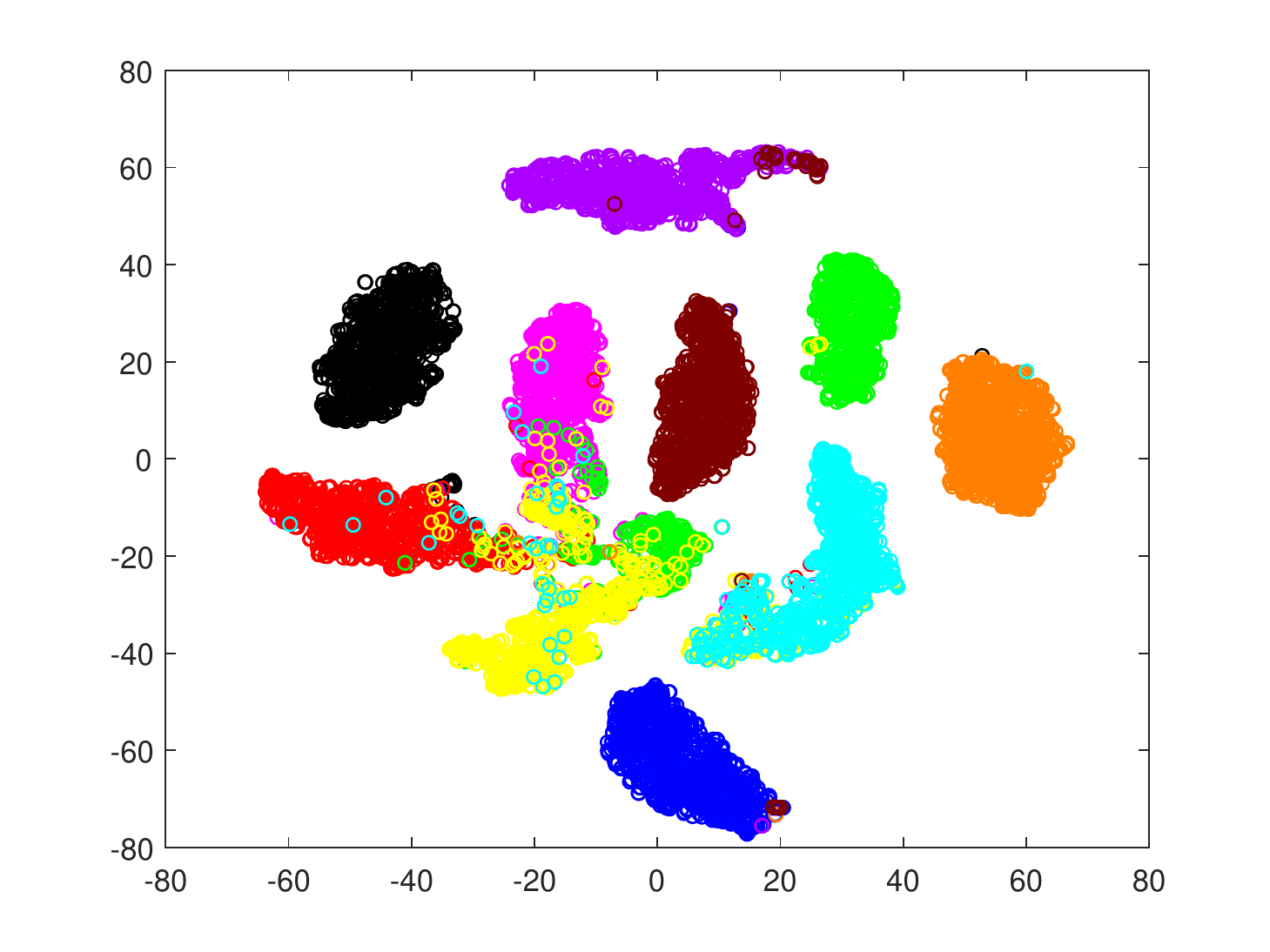}}
	\caption{The t-SNE representations of real-valued features learned by different methods: (a) DPSH, (b) DAPH, (c) DSDH, (d) DDSH, (e) ADSH, (f) DAGH, (g) the proposed DSAH-1, and (h) the proposed DSAH-2.}  	\label{tSNE}
\end{figure*}

\subsection{Visualization}
Figure \ref{tSNE} shows the visualization of t-SNE representation \cite{24} learned by all deep hashing methods on the Fashion-MNIST data set. In Figure \ref{tSNE}, each color represents one class of samples. Compared with other methods, the proposed DSAH-1 and DSAH-2 can better characterize the intra-class compactness and the inter-class separability of data, which further demonstrates the effectiveness of the proposed framework. 

\begin{table*}
	\footnotesize
	\centering
	\setlength{\tabcolsep}{1.2mm}
	\caption{Ablation study of the proposed DSAH in MAP (\%) on all used data sets.}
	\begin{tabular}{l|cccc|cccc|cccc}
		\hline
		\hline
		\multirow{2}{*}{Method}&\multicolumn{4}{c|}{CIFAR-10}   &\multicolumn{4}{c|}{Fashion-MNIST} &\multicolumn{4}{c}{NUS-WIDE}   \\
		\cline{2-13}
		&12 &24 &32 &48  &12 &24 &32 &48  &12 &24 &32 &48   \\
		\hline
		DSAH-1-A&93.72   &94.78    &94.18     &94.70    
		&91.53  &91.02  &92.06   &92.44 &74.85	&76.60	&76.11	&76.38
		\\
		DSAH-1-B&\textbf{94.26}   &94.24    &94.57     &94.46    
		&91.49  &91.84  &92.59   &92.44   & 72.96	&73.53	&77.42	&77.63
		\\
		DSAH-1-C&93.76   &94.20    &94.06     & 94.49   
		&91.20   &91.80   &92.32    &92.34   &71.67	&72.28	&73.56	&73.44
		\\			
		DSAH-1-D&93.70   &94.70    &94.09     &94.31    
		&89.81  &91.46  &92.21   &91.21   &61.03	&63.57	&67.64	&69.55  
		\\
		DSAH-1&{94.24}  &\textbf{95.26}   &\textbf{95.25}  &\textbf{95.16}  
		&\textbf{91.86}  &\textbf{92.20} &\textbf{92.90}   &\textbf{92.90}    &\textbf{76.96}	&\textbf{81.03}	&\textbf{81.64}	&\textbf{82.80}
		\\
		\hline
		DSAH-2-A &93.67    &94.56    &94.37    &94.68    &91.09   &91.55    &91.79    &91.48   &72.53	&74.60	&75.79	&75.95
		\\
		DSAH-2-B &\textbf{93.80}    &94.77    &94.44    &94.64   &91.56    &91.92    &92.56   &92.76    &74.67	&77.42	&76.28	&75.33  \\
		DSAH-2-C &93.74    &94.61    &94.47    &94.75    &90.54   &91.52    &91.92    &\textbf{92.83}    &71.70	&72.58	&72.21	&72.94\\
		DSAH-2-D &93.46    &94.04    &94.51    &94.34    &91.25   &90.93    &90.74    &92.51   &63.49	&61.36	&68.54	&70.07	\\	
		DSAH-2	
		&93.77   &\textbf{94.94}   &\textbf{95.02}   &\textbf{95.10}  
		&\textbf{91.70}   &\textbf{92.09}   &\textbf{93.12}   & 92.68  
		&\textbf{78.47}   &\textbf{80.36}   &\textbf{80.89}   &\textbf{82.70}\\
		\hline
		\hline
	\end{tabular}
	\label{ablation}
\end{table*}

\subsection{Ablation Study}
We investigate the retrieval performance for four variants of the proposed DSAH-1 and DSAH-2. The variant A removes the dual semantic regression loss $\mathcal{R}$. The variant B only utilizes single label $\mathbf{Y}$ and the variant C only uses single label $\mathbf{R}$ in the dual semantic regression loss $\mathcal{R}$. And the variant D ignores balance constraint $\mathbf{H}^T\mathbf{1}=0$ in the process of binary code learning. Table \ref{ablation} lists the MAP results of these variants on all used data sets. As shown in Table \ref{ablation}, the proposed DSAH-1 and DSAH-2 achieve the best retrieval performance in most cases on single-label CIFAR-10 and Fashion-MNIST data sets. On multi-label NUS-WIDE data set, the performance differences between the proposed methods and their four variants are significant, which verifies the effectiveness of each component of the proposed DSAH.

\section{Conclusion}
This paper proposes a novel deep hashing framework called DSAH, 
which consists of three components including dual semantic regression, pairwise similarity preserving, and class structure quantization. The proposed DSAH adopts a distance-similarity product function for data similarity preserving and incorporates class structure prior into the quantization process, thereby providing discriminative feedback for network training. The dual semantic labels are designed to characterize the intra-class and inter-class information of data, so as to learn semantic-sensitive binary codes. Extensive experiments on three well-known data sets demonstrate the promising network generalization performance of the proposed method. In the future, we will use neural network to replace linear projection for information integration of dual semantic labels.

\section*{Acknowledgments} 
This work was supported in part by the National Natural Science Foundation of China (No. 62076164), in part by the Guangdong Basic and Applied Basic Research Foundation (No. 2021A1515011318, 2021A1515011861), and in part by Shenzhen Science and Technology Program (No. JCYJ20210324094601005). 

~\\


\leftline{ {\bf References}}

\begin{thebibliography}{51}
	
\bibitem{3}
Z. Cao, M. Long, J. Wang, P.S. Yu, HashNet: deep learning to hash by continuation, in: Proceedings of the IEEE conference on computer vision and pattern recognition, 2017, pp. 5609-5618. 

\bibitem{4}
K. Chatfield, K. Simonyan, A. Vedaldi, A. Zisserman, Return of the devil in the details: delving deep into convolutional nets, 2014, arXiv preprint arXiv:1405.3531.

\bibitem{5}
W. Chen, J.T. Wilson, S. Tyree, K.Q. Weinberger, Y. Chen, Compressing neural networks with the hashing trick, in: Proceeding of the International Conference on Machine Learning, 2015, pp. 2285-2294.

\bibitem{6}
Y. Chen, Z. Lai, Y. Ding, K. Lin, W. Wong, Deep supervised hashing with anchor graph, in: Proceedings of the IEEE/CVF international conference on computer vision, 2019, pp. 9795-9803. 

\bibitem{7}
C. Da, G. Meng, S. Xiang, K. Ding, S. Xu, Q. Yang, C. Pan, Nonlinear asymmetric multi-valued hashing, IEEE Trans. Pattern Anal. Mach. Intell. 41 (2019) 2660-2676.
 
\bibitem{8}
M. Datar, N. Immorlica, P. Indyk, V.S. Mirrokni, Locality-sensitive hashing scheme based on p-stable distributions, in: Proceedings of the International Symposium on Computational Geometry, 2004, pp. 253-262.

\bibitem{9}
W. Dong, M. Charikar, K. Li, Asymmetric distance estimation with sketches for similarity search in high-dimensional spaces, in: Proceedings of the Annual International ACM SIGIR Conference on Research and Development in Information Retrieval, 2008, pp. 123-130.

\bibitem{10}
Y. Gong, S. Lazebnik, A. Gordo, F. Perronnin, Iterative quantization: a procrustean approach to learning binary codes for large-scale image retrieval, {IEEE} Trans. Pattern Anal. Mach. Intell. 35 (2013) 2916-2929.

\bibitem{11}
A. Gordo, F. Perronnin, Y. Gong, S. Lazebnik, Asymmetric distances for binary embeddings, IEEE Trans. Pattern Anal. Mach. Intell. 36 (2014) 33-47. 


\bibitem{13}
J. Gui, T. Liu, Z. Sun, D. Tao, T. Tan, Fast supervised discrete hashing, IEEE Trans. Pattern Anal. Mach. Intell. 40 (2018) 490-496.

\bibitem{14}
J. Gui, T. Liu, Z. Sun, D. Tao, T. Tan, Supervised discrete hashing with relaxation, IEEE Trans. Neural Networks Learn. Syst. 29 (2018) 608-617.

\bibitem{15}
D. Hu, F. Nie, X. Li, Discrete spectral hashing for efficient similarity retrieval, IEEE Trans. Image Process. 28 (2019) 1080-1091. 

\bibitem{16}
P. Hu, X. Wang, L. Zhen, D. Peng, Separated variational hashing networks for cross-modal retrieval, in: Proc. 27th {ACM} Int. Conf. Multimed., 2019: pp. 1721-1729.


\bibitem{18}
M. Jagadeesan, Understanding sparse JL for feature hashing, in: Proceedings of the Conference on Neural Information Processing Systems, 2019, pp. 15203-15213.

\bibitem{19}
Q.-Y. Jiang, X. Cui, W.-J. Li, Deep discrete supervised hashing, IEEE Trans. Image Process. 27 (2018) 5996-6009.

\bibitem{20}
Q.-Y. Jiang, W.-J. Li, Asymmetric deep supervised hashing, in: Proceedings of AAAI Conference on Artificial Intelligence, (2017) 3342-3349.

\bibitem{21}
B. Kulis, T. Darrell, Learning to hash with binary reconstructive embeddings, in: Proceedings of the Conference on Neural Information Processing Systems, 2009, pp. 1042-1050.

 
\bibitem{24}  
V.D.M. Laurens, G. Hinton, Visualizing data using t-SNE, J. Mach. Learn. Res. 9 (2008) 2579-2605.

\bibitem{25}
C.-X. Li, T.-K. Yan, X. Luo, L. Nie, X.-S. Xu, Supervised robust discrete multimodal hashing for cross-media retrieval, {IEEE} Trans. Multim. 21 (2019) 2863-2877.

\bibitem{26}
N. Li, C. Li, C. Deng, X. Liu, X. Gao, Deep joint semantic-embedding hashing, in: Proceedings of the International Joint Conferences on Artificial Intelligence, 2018, pp. 2397-2403. 

\bibitem{27}
Q. Li, Z. Sun, R. He, T. Tan, A general framework for deep supervised discrete hashing, Int. J. Comput. Vis. 128 (2020) 2204-2222. 

\bibitem{28}
W. Li, S. Wang, W. Kang, Feature learning based deep supervised hashing with pairwise labels, in: Proceedings of the International Joint Conferences on Artificial Intelligence, 2016, pp. 1711-1717.


\bibitem{30}
Y. Li, R. Xiao, X. Wei, H. Liu, S. Zhang, X. Du, GLDH: toward more efficient global low-density locality-sensitive hashing for high dimensions, Inf. Sci. 533 (2020) 43-59. 

\bibitem{31}
C. Liu, G. Yu, C. Chang, H. Rai, J. Ma, S.K. Gorti, M. Volkovs, Guided similarity separation for image retrieval, in: Proceedings of the Conference on Neural Information Processing Systems, 2019, pp. 1556-1566.

\bibitem{32}
H. Liu, R. Wang, S. Shan, X. Chen, Deep supervised hashing for fast image retrieval, in: Proceedings of the IEEE conference on computer vision and pattern recognition, 2016, pp. 2064-2072.


\bibitem{34}
W. Liu, C. Mu, S. Kumar, S.-F. Chang, Discrete graph hashing, in: Proceedings of the Conference on Neural Information Processing Systems, 2014, pp. 3419-3427.

\bibitem{35}
W. Liu, J. Wang, S. Kumar, S.-F. Chang, Hashing with graphs, in: Proceedings of the International Conference on Machine Learning, 2011, pp. 1-8.


\bibitem{37}
X. Lu, L. Zhu, J. Li, H. Zhang, H.T. Shen, Efficient supervised discrete multi-view hashing for large-scale multimedia search, IEEE Trans. Multimed. 22 (2020) 2048-2060. 

\bibitem{38}
Y. Luo, Z. Huang, Y. Li, F. Shen, Y. Yang, P. Cui, Collaborative learning for extremely low bit asymmetric hashing, IEEE Trans. Knowl. Data Eng. (2020) doi: 10.1109/TKDE.2020.2977633. 

\bibitem{39}
B. Neyshabur, N. Srebro, R.R. Salakhutdinov, Y. Makarychev, P. Yadollahpour, The power of asymmetry in binary hashing, in: Proceedings of the Conference on Neural Information Processing Systems, 2013, pp. 2823-2831.

\bibitem{40}
F. Shen, X. Gao, L. Liu, Y. Yang, H.T. Shen, Deep asymmetric pairwise hashing, in: Proceedings of the ACM International Conference on Multimedia, 2017, pp. 1522-1530.

\bibitem{41}
F. Shen, C. Shen, W. Liu, H.T. Shen, Supervised discrete hashing, in: Proceedings of the IEEE conference on computer vision and pattern recognition, 2015, pp. 37-45. 

\bibitem{42}
F. Shen, C. Shen, Q. Shi, A. van den Hengel, Z. Tang, Inductive hashing on manifolds, in: Proceedings of the IEEE conference on computer vision and pattern recognition, 2013, pp. 1562-1569.

\bibitem{43}
Y. Shen, Y. Feng, B. Fang, M. Zhou, S. Kwong, B. Qiang, DSRPH: deep semantic-aware ranking preserving hashing for efficient multi-label image retrieval, Inf. Sci. 539 (2020) 145-156.


\bibitem{46}
A. Vedaldi, K. Lenc, MatConvNet: convolutional neural networks for MATLAB, in: Proceedings of the ACM International Conference on Multimedia, 2015, pp. 689-692.

\bibitem{47}
H. Venkateswara, J. Eusebio, S. Chakraborty, S. Panchanathan, Deep hashing network for unsupervised domain adaptation, in: Proceedings of the IEEE conference on computer vision and pattern recognition, 2017, pp. 5385-5394.

\bibitem{48}
J. Wang, T. Zhang, J. Song, N. Sebe, H.T. Shen, A survey on learning to hash, IEEE Trans. Pattern Anal. Mach. Intell. 40 (2018) 769-790. 

\bibitem{49}
W. Wang, H. Zhang, Z. Zhang, L. Liu, L. Shao, Sparse graph based self-supervised hashing for scalable image retrieval, Inf. Sci. 547 (2021) 622-640. 


\bibitem{51}
Y. Weiss, A. Torralba, R. Fergus, Spectral hashing, in: Proceedings of the Conference on Neural Information Processing Systems, 2008, pp. 1753-1760.

\bibitem{52}
Y. Xia, K. He, P. Kohli, J. Sun, Sparse projections for high-dimensional binary codes, in: Proceedings of the IEEE conference on computer vision and pattern recognition, 2015, pp. 3332-3339.


\bibitem{54}
P. Xu, Y. Huang, T. Yuan, K. Pang, Y. Song, T. Xiang, T.M. Hospedales, Z. Ma, J. Guo, SketchMate: deep hashing for million-scale human sketch retrieval, in: Proceedings of the IEEE conference on computer vision and pattern recognition, 2018, pp. 8090-8098.

\bibitem{55}
X. Yan, L. Zhang, W.-J. Li, Semi-supervised deep hashing with a bipartite graph, in: Proceedings of the International Joint Conference on Artificial Intelligence, 2017, pp. 3238-3244.


\bibitem{59}
P. Zhang, W. Zhang, W.-J. Li, M. Guo, Supervised hashing with latent factor models, in: Annual International ACM SIGIR Conference on Research and Development in Information Retrieval, 2014, pp. 173-182.



\end{thebibliography}
\end{document}